\documentclass{article}

\usepackage[preprint]{neurips_2024}

\usepackage[utf8]{inputenc} 
\usepackage[T1]{fontenc}    
\usepackage[breaklinks,bookmarksdepth=3,bookmarksopen=true,bookmarksopenlevel=2,bookmarksnumbered=true]{hyperref}
\usepackage{url}            
\usepackage{booktabs}       
\usepackage{amsfonts}       
\usepackage{nicefrac}       
\usepackage{microtype}      
\usepackage{xcolor}         

\usepackage[inline]{enumitem} 
\usepackage{color}
\usepackage{multirow}
\usepackage[makeroom]{cancel}
\usepackage[accsupp]{axessibility}
\usepackage{graphicx}
\usepackage{subcaption}
\usepackage{amsmath}
\usepackage{bm}
\usepackage{algorithm}
\usepackage{listings}
\usepackage{wrapfig}
\usepackage{floatflt}
\usepackage{blindtext}

\definecolor{turquoise}{cmyk}{0.65,0,0.1,0.3}
\definecolor{purple}{rgb}{0.65,0,0.65}
\definecolor{dark_purple}{rgb}{0.5,0,0.5}
\definecolor{dark_green}{rgb}{0, 0.5, 0}
\definecolor{orange}{rgb}{0.8, 0.6, 0.2}
\definecolor{red}{rgb}{0.8, 0.2, 0.2}
\definecolor{darkred}{rgb}{0.6, 0.1, 0.05}
\definecolor{blueish}{rgb}{0.0, 0.3, .6}
\definecolor{light_gray}{rgb}{0.7, 0.7, .7}
\definecolor{pink}{rgb}{1, 0, 1}
\definecolor{greyblue}{rgb}{0.25, 0.25, 1}

\usepackage[capitalise,noabbrev]{cleveref}
\Crefname{section}{Sec.}{Secs.}

\usepackage{etoolbox}
\makeatletter
\AfterEndEnvironment{algorithm}{\let\@algcomment\relax}
\AtEndEnvironment{algorithm}{\kern2pt\hrule\relax\vskip3pt\@algcomment}
\let\@algcomment\relax
\newcommand\algcomment[1]{\def\@algcomment{\footnotesize#1}}
\renewcommand\fs@ruled{\def\@fs@cfont{\bfseries}\let\@fs@capt\floatc@ruled
  \def\@fs@pre{\hrule height.8pt depth0pt \kern2pt}%
  \def\@fs@post{}%
  \def\@fs@mid{\kern2pt\hrule\kern2pt}%
  \let\@fs@iftopcapt\iftrue}
\makeatother

\newcommand{\ie}{\emph{i.e.}}
\newcommand{\eg}{\emph{e.g.}}

\newcommand*\circled[1]{\textcircled{\scalebox{0.8}{\textbf{#1}}}}

\newcommand{\Ours}{\textsc{EG4D}}

\title{\Ours{}: Explicit Generation of 4D Object \\
without Score Distillation
}

\author{%
Qi Sun$^{1, 2}\thanks{Equal contribution.}$
\quad 
Zhiyang Guo$^{1 *}$
\quad 
Ziyu Wan$^{2}$
\quad 
Jing Nathan Yan$^{3}$
\\
\textbf{Shengming Yin$^{1}$}
\quad 
\textbf{Wengang Zhou}$^{1}$
\quad 
\textbf{Jing Liao}$^{2}$ 
\quad 
\textbf{Houqiang Li}$^{1}$ \\
  $^{1}$USTC
  $^{2}$City University of Hong Kong, $^{3}$Cornell University\\
  \texttt{\{qisun, guozhiyang, shengmingyin\}@mail.ustc.edu.cn}, \texttt{raywzy@gmail.com}
  \\
  \texttt{jy858@cornell.edu}, \texttt{jingliao@cityu.edu.hk}, \texttt{\{zhwg, lihq\}@ustc.edu.cn}\\
}

\begin{document}

\maketitle

\begin{abstract}
  In recent years, the increasing demand for dynamic 3D assets in design and gaming applications has given rise to powerful generative pipelines capable of synthesizing high-quality 4D objects.
  Previous methods generally rely on score distillation sampling (SDS) algorithm to infer the unseen views and motion of 4D objects, thus leading to unsatisfactory results with defects like over-saturation and Janus problem.
  Therefore, inspired by recent progress of video diffusion models, we propose to optimize a 4D representation by explicitly generating multi-view videos from one input image.
  However, it is far from trivial to handle practical challenges faced by such a pipeline, including dramatic temporal inconsistency, inter-frame geometry and texture diversity, and semantic defects brought by video generation results.
  To address these issues, we propose \Ours{}, a novel multi-stage framework that generates high-quality and consistent 4D assets without score distillation.
  Specifically, collaborative techniques and solutions are developed, including an attention injection strategy to synthesize temporal-consistent multi-view videos, a robust and efficient dynamic reconstruction method based on Gaussian Splatting, and a refinement stage with diffusion prior for semantic restoration.
  The qualitative results and user preference study demonstrate that our framework outperforms the baselines in generation quality by a considerable margin.
  Code will be released at \url{https://github.com/jasongzy/EG4D}.
\end{abstract}

\section{Introduction}

Recent years have seen a surge in the development of generative models capable of producing intelligible text~\cite{Brown2020NeurIPS, Touvron2023ARXIV, OpenATI2023ARXIV}, photo-realistic images~\cite{Ramesh2021ICML, Rombach2022CVPR, Rombach2022CVPR, Sauer2023ICML}, video sequences~\cite{Skorokhodov2022CVPR, Bahmani2022ARXIV, Singer2022ARXIV}, 3D~\cite{chan2022efficient, poole2022dreamfusion, Lin2023CVPR, tang2023dreamgaussian} and 4D (dynamic 3D) assets~\cite{ren2023dreamgaussian4d, zhao2023animate124, jiang2023consistent4d, singer2023text}.
Particularly with 4D assets, manual creation is a laborious task that requires considerable expertise from highly skilled designers. Systems capable of automatically generating realistic and diverse 4D content could greatly streamline the workflows of artists and designers, potentially unlocking new realms of creativity through ``generative art''~\cite{Bailey2020ArtInAmerica}. 

Due to the scarcity of open-sourced annotated multi-view dynamic data, 
previous works~\cite{yin20234dgen, xu2024comp4d, bahmani2024tc4d, bahmani20234d, ren2023dreamgaussian4d, zhao2023animate124, jiang2023consistent4d, singer2023text, zheng2024unified, ling2023align} rely on the score distillation sampling (SDS)~\cite{poole2022dreamfusion} or its variants from pre-trained 2D diffusion models to distill information about unseen views and motion of objects.
Despite the impressive performance, their rendering results still suffer from highly saturated texture~\cite{wang2023prolificdreamer} and multi-face geometry (Janus problem)~\cite{armandpour2023re}, thus leading to less photo-realistic generations.

Motivated by recent progress in video diffusion models~\cite{voleti2024sv3d, blattmann2023stable, brooks2024video}, we propose a novel multi-stage framework, \textbf{\Ours{}}, for \textbf{E}xplicitly \textbf{G}enerating \textbf{4D} videos and then reconstructing 4D assets from them.
\Ours{} goes beyond simply adapting video generation results, as the synthesized frames inevitably suffer from temporal inconsistency and limited visual quality. More specifically, in the vanilla ``frame-by-frame'' reconstruction, the independence and diversity of multi-view diffusion will cause appearance inconsistency across different timestamps, particularly in unseen views.

To address these challenges, we first design an attention injection mechanism, allowing each multi-view diffusion inference to perceive temporal information through cross-frame latent exponential moving average (EMA). This training-free strategy effectively alleviates the inconsistency issue at the video level and ensures high-quality training samples for optimizing the following 4D representation. In the next stage of 4D reconstruction, we choose 4D Gaussian Splatting (4D-GS)~\cite{4dgs} as our representation to take advantage of its efficient training and rendering capability.

Moreover, existing GS-based dynamic reconstruction methods~\cite{4dgs,yang2023deformable,huang2023sc} commonly assume that appearance variations between different timestamps are caused by the geometric deformation of Gaussian splats. However, this assumption does not hold since unwanted color variations of texture details still exist in our synthesized images produced by video diffusions, even with the proposed attention injection strategy. We manage to disentangle such detailed texture inconsistencies from desired geometric deformation by introducing an extra color transformation network, enabling texture-consistent 4D rendering.
Furthermore, we leverage image-to-image diffusion models to refine the rendered images and fine-tune our 4D representation, achieving better generation quality.

The qualitative results and human preferences validate that our \Ours{} outperforms SDS-based baselines by a large margin, producing 4D content with realistic 3D appearance, high image fidelity, and fine temporal consistency.
Extensive ablation studies also showcase our effective solutions to the challenges in reconstructing 4D representation with synthesized videos.

\section{Related Works}

In this section, we present the recent progress of video diffusion models and 4D generation.
More discussion on related works can be found in \cref{app:related}.

\noindent\textbf{Video diffusion models.}
Diffusion models~\cite{hoDenoisingDiffusionProbabilistic2020}, characterized by their superior generative capabilities, have become dominant in the field of video generation ~\cite{hoVideoDiffusionModels2022,singerMakeAVideoTexttoVideoGeneration2022,hoImagenVideoHigh2022,blattmann2023stable,yinNUWAXLDiffusionDiffusion2023}. Among them, VDM~\cite{hoVideoDiffusionModels2022} replaces the typical 2D U-Net for modeling images with a 3D U-Net. Make-A-Video~\cite{singerMakeAVideoTexttoVideoGeneration2022} successfully extends a diffusion-based T2I model to T2V without text-video pairs. 
Text2Video-Zero~\cite{khachatryanText2VideoZeroTexttoImageDiffusion2023} achieve zero-shot text-to-video generation using only a pre-trained text-to-image diffusion model without any further fine-tuning or optimization. Following Latent Diffusion Models~\cite{rombachHighResolutionImageSynthesis2022}, Video-LDM~\cite{blattmannAlignYourLatents2023} and AnimateDiff~\cite{guoAnimateDiffAnimateYour2023c} introduce additional temporal layers designed to model the temporal consistency. Stable Video Diffusion~\cite{blattmann2023stable}, trained on well-curated high quality video dataset, presents robust text-to-video and image-to-video generation capabilities across various domains. Recently, SV3D~\cite{voleti2024sv3d} adapts image-to-video generation for novel view synthesis by leveraging the generalization and multi-view consistency of the video models.
Different from these works, we aim to explicitly generate 4D videos with both temporal and multi-view consistency using two orthogonal video diffusion models.


\noindent\textbf{4D generation.}
Following the line of text-to-3D synthesis~\cite{poole2022dreamfusion, wang2023prolificdreamer, wan2023cad}, one line of research explores the text-conditioned 4D generation~\cite{yin20234dgen, cai2023generative, xu2024comp4d, bahmani2024tc4d, zheng2024unified, bahmani20234d, singer2023text, zheng2024unified, ling2023align}. They use score distillation sampling (SDS)~\cite{poole2022dreamfusion} to optimize the 4D representations, like KPlanes~\cite{fridovich2023k}, Hexplanes~\cite{cao2023hexplane} and Deformable Gaussians~\cite{yang2023deformable}.
MAV3D~\cite{singer2023text} employs temporal SDS to transfer the motion from text-to-video diffusions~\cite{singer2022make}) to a dynamic NeRF.
4D-fy~\cite{bahmani20234d} exploits hybrid score distillation methods by alternating optimization procedure to improve the structure and quality of the 4D model.
AYG~\cite{ling2023align} explores compositional 4D generation with 3D Gaussian Splatting.
Inspired by recent advancement in image-to-3D models~\cite{liu2023zero, liu2023one, shi2023MVDream}, several works~\cite{zhao2023animate124, ren2023dreamgaussian4d, jiang2023consistent4d} explore the field of image/video-conditioned 4D generation.
Animate124~\cite{zhao2023animate124} pioneers on this task in a coarse-to-fine fashion: it first optimizes deformation with multi-view diffusions, then corrects the details with ControlNet~\cite{zhang2023adding}.
DreamGaussian4D~\cite{ren2023dreamgaussian4d}  adopts explicit modeling of spatial transformations in Gaussian Splatting, achieving minute-level generation.
Although score distillation algorithm can infer motion and unseen views from 2D diffusion models, it suffers from some imperfections like over-saturation and Janus problem. 
Our framework gets around these problems by explicitly generating multi-views of dynamic object, and then uses them to reconstruct 4D representations.

\section{Preliminaries}

\noindent\textbf{Video diffusion.}
In this work, we use two different video diffusion models: Stable Video Diffusion~\cite{blattmann2023stable} (SVD) and SV3D~\cite{voleti2024sv3d}. SVD generates a sequence of video frames $\{I_t | t\in \{0,\cdots, T\}\}$ conditioned on an initial image $I_0$ or text prompt.
SV3D is a pose-conditioned image-to-multiviews model that takes a reference image $I_0$ and a series of camera poses $\{c_p| p\in \{1,\cdots, N\}\}$, producing a sequence of video frames $\{I_{p}| p\in \{1,\cdots, N\}\}$ corresponding to the specified pose (camera parameters) sequence.
Both SVD and SV3D adopt similar video diffusion architecture~\cite{ling2023align} with spatial and temporal attention layers.

\noindent\textbf{3D Gaussian Splatting.}
3DGS~\cite{3dgs} is an explicit representation using millions of 3D Gaussians to model a scene. Each Gaussian is characterized by a set of learnable parameters as follows:
\textbf{1)} 3D center;
\textbf{2)} 3D rotation;
\textbf{3)} 3D size (scaling factor);
\textbf{4)} view-dependent RGB color represented by spherical harmonics coefficients (with degrees of freedom $k$): $\bm{h} \in \mathbb{R}^{3(k+1)^2} \rightarrow \bm{c} \in \mathbb{R}^3$;
\textbf{5)} opacity.
%
Here a color decoder $\Phi^{sh}$ is used to turn the spherical harmonics coefficients $\bm{h}$ and the view direction $\bm{\gamma}$ into an actual RGB color $\bm{c}$.
For a position
in the scene, 
each
Gaussian makes its contribution at that coordinate according to the standard Gaussian function weighted by its opacity.
The differentiable rendering of 3DGS applies the splatting techniques~\cite{3dgs}.
For a certain pixel, the point-based rendering computes its color by evaluating the blending of depth-ordered points overlapping that pixel via the volume rendering equation~\cite{volumnrender}.
The optimization of Gaussian parameters is then supervised by the reconstruction loss (difference between rendered and ground-truth images).

\begin{figure}[t]
    \centering
    \includegraphics[width=\linewidth]{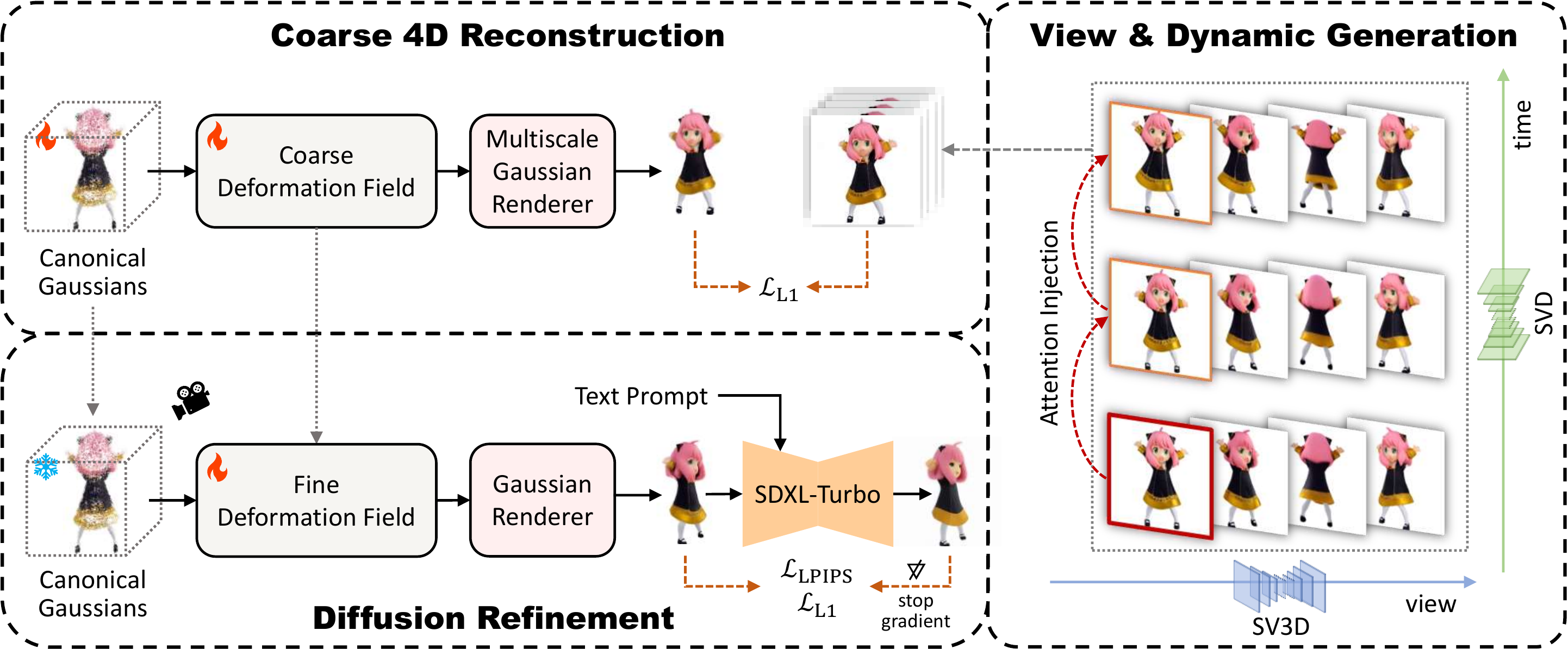}
    \caption{\textbf{Framework of \Ours{}.} In video generation (right, \cref{subsec:videogen}), we use SVD to produce dynamic frames, and then use SV3D equipped with attention injection to generate temporal-consistent multi-view images.
    In coarse 4D reconstruction (left top, \cref{subsec:recon}), we optimize the 4D Gaussian Splatting with additional color affine transformation with the annotated multi-view images produced by Stage I.
    In diffusion refinement (left bottom, \cref{subsec:refine}), we freeze the canonical Gaussians and further fine-tune the temporal deformation network with images refined by an image-to-image diffusion model.}
    \label{fig:framework}
\end{figure}

\section{4D Object Generation}
Given an object image, we want to generate the 4D representation of it, enabling free-view dynamic rendering.
To this end, we introduce a multi-stage framework {(generation-reconstruction-refinement)} for 4D object generation, as illustrated in \cref{fig:framework}. Network details can be found in \cref{app:network}.

\subsection{Stage I: View and Dynamic Generation with Video Diffusions}
\label{subsec:videogen}

In this stage, we employ two orthogonal video diffusion models to generate samples for the later 4D representation optimization. Given a reference image, we use SVD~\cite{blattmann2023stable} to generate a sequence of video frames $\{I_t | t\in \{0,\cdots, T\}\}$, where $t$ is the timestamp.
Next, we utilize SV3D~\cite{voleti2024sv3d} to generate multi-view images $\{I_{t,p} | p \in \{1, \cdots, N\}\}$ with a predefined camera pose sequence for each frame $I_t$.
However, vanilla ``frame-by-frame'' reconstruction causes significant \emph{temporal differences} due to the diverse nature of SV3D inferences for those frames.
Hence, we hope to exploit temporal context to guide the otherwise independent generating process, thereby obtaining results that are as temporally consistent as possible.
To this end, we introduce the training-free \emph{attention injection} strategy during our SV3D inference.

Specifically, in each self-attention module of the spatial layers of a diffusion UNet, we simultaneously consider the visual information from the current reference frame and the frames at previous timestamps, and implement the attention injection by \emph{spatial KV latent blending}
formulated as

\vspace{-3mm}
\begin{equation}
{\bm{z}_t} \leftarrow \alpha \bm{z}^*_{t} + (1 - \alpha) \bm{z}_{t-1}, 
\end{equation}
\vspace{-1mm}
\begin{equation}
    \bm{Q} = \bm{W}^q  {\bm{z}}^*_t,  \bm{K} =  \bm{W}^k {\bm{z}_t}, \bm{V} = \bm{W}^v {\bm{z}_t},
\end{equation}
\vspace{-2mm}
\begin{equation}
    \text{Attention} (\bm{Q}, \bm{K}, \bm{V}) = \text{Softmax}(\frac{\bm{Q}  \bm{K}^T}{\sqrt{d_k}} \bm{V}),
\end{equation}

where $\bm{z}_t$ is the exponential moving average (EMA) of the current multi-view latent ${z_t}^*$ and the one from the previous timestamp $z_{t-1}$, with the blending weight $\alpha$. $d_k$ is the key dimension.

\subsection{Stage II: Coarse Reconstruction with Gaussian Splatting}
\label{subsec:recon}

With the synthesized multi-view images $\{I_{t,p} | t\in \{0,\cdots, T\}, p\in  \{1, \cdots, N\} \}$ of the dynamic object, we optimize a 4D representation of it to enable free-viewpoint rendering.
It is worth noting that in this stage, our objective is \emph{not} simply reconstructing an object according to multi-view observations.
Although the design in \cref{subsec:videogen} significantly alleviates the temporal inconsistency problem, those synthesized ``ground-truth'' images still suffer from varying degrees of inconsistency in color details.
Therefore, we propose to optimize a 4D representation based on 3D Gaussian Splatting~\cite{3dgs} with additional insights into the robustness against texture inconsistencies and semantic defects.

\paragraph{Canonical Gaussians \& deformation field.}
Considering both performance and efficiency, we build our 4D representation upon 4D Gaussian Splatting (4D-GS)~\cite{4dgs}.
4D-GS utilizes a deformation field to predict each Gaussian's geometric offsets at a given timestamp relative to a mean canonical state. 
This deformation field is composed of a multi-resolution HexPlane~\cite{cao2023hexplane} and MLP-based decoders. 
For each Gaussian at a certain timestamp, the model queries the Hexplane with a 4D coordinate ($x\text{-}y\text{-}z\text{-}t$) and decodes the obtained feature $\bm{f}_t$ into the position, rotation, and scaling deformation values. The entire dynamic scene is then jointly reconstructed by optimizing both canonical Gaussians and the deformation field, enabling implicit global interactions of visual information.

\paragraph{Color transformation against texture inconsistency.}
While vanilla 4D-GS is theoretically able to model temporal inconsistencies through per-frame geometric deformation of Gaussians, it is hard to optimize and leads to significant redundancy in Gaussian quantity~\cite{guo2024motion}. 
Even if all the inconsistencies are faithfully reconstructed, these unnatural variations in texture details across time will result in significant degradation of visual performance.
To address this problem, we want to disentangle such detailed texture inconsistencies from geometric deformation. Those temporal differences can still be modeled as per-timestamp states, while one of them can be manually selected to dominate the final temporal-consistent rendering.
We choose a simple but effective way that performs time-specific color transformation.
Formally, a new color decoder denoted by $\Phi^{c}$ is introduced as follows:

\begin{equation}
     \bm{c} = \Phi^{c}(\bm{h},\bm{\gamma}) = \bm{W}^c_t \Phi^{sh}(\bm{h},\bm{\gamma}) + \bm{b}^c_t,
\end{equation}
\begin{equation}
     \bm{W}^c_t, \bm{b}^c_t = \text{MLP}(\bm{f}_t),
\end{equation}

where $\bm{h}$ is the spherical harmonics coefficients of Gaussians, $\bm{\gamma}$ is the view direction, and $\Phi^{sh}$ is the spherical harmonic decoder. $\bm{W}^c_t$ and $\bm{b^c_t}$ are weights and bias predicted by an extra MLP-based color head from per-Gaussian time-specific feature $\bm{f}_t$ from the HexPlane.
Such kind of affine transformation is competent in modeling texture inconsistencies caused by ambient occlusion and other factors~\cite{li2022tava,darmon2024robust}. 
During 4D rendering at test time, we take one of the timestamps, \eg, the first frame, as the reference time and use the corresponding feature $\bm{f}_0$ to get the Gaussian colors, thereby rendering texture-consistent 4D assets.

\paragraph{Multiscale rendering augmentation.}

Generally, for a reconstruction task, supervision with high-resolution ground-truth images can provide more information about high-frequency details and benefit the rendering quality. 
However, in our task, those synthesized images often have high-frequency noises at specific views or timestamps.
Training with them leads to meaningless view- and frame-overfitting and adds more burden to later refinement.
To address this issue, we propose a multiscale augmentation strategy. During optimization, we randomly downsample the ground-truth images within a reasonable ratio range. The rendering parameters of the Gaussian rasterizer are modified accordingly, enabling multiscale supervision with the reconstruction loss.

\subsection{Stage III: Refinement with Diffusion Priors}
\label{subsec:refine}

\setlength{\intextsep}{0pt}
\begin{wrapfigure}{r}{0.47\textwidth}
  \centering
  \vspace{-3mm}
  \resizebox{\linewidth}{!}{
  \includegraphics{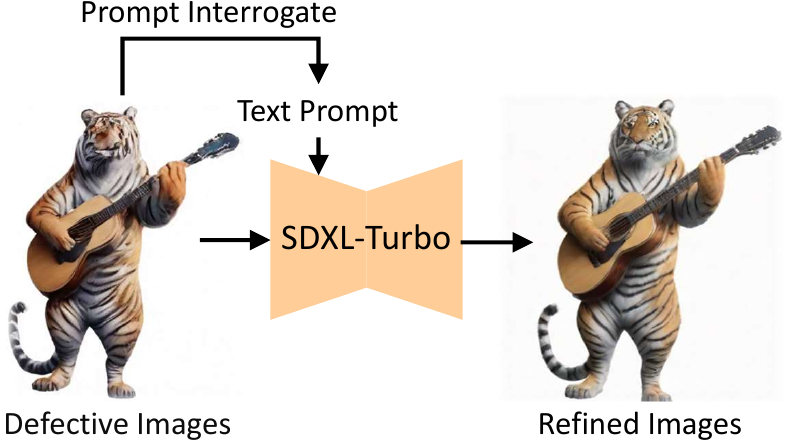}
  }
  \caption{\textbf{Illustration of diffusion refinement.} 
  }
  \label{fig:refine}
\end{wrapfigure}

Videos produced by diffusion models often suffer from semantic defects (\cref{fig:refine} left) and motion blur.
Fortunately, image-to-image diffusion models provide a strong prior to refine the semantic details while preserving object identity and style. 
We leverage these diffusion-refined images (\cref{fig:refine} right) to fine-tune our 4D representation further.
Specifically, we first render an image $I_{t,p}$ at the timestamp $t$ and camera pose $p$.
Then we encode the image $I_{t,p}$ into a VAE latent $w$, add noise to the latent, and feed it into the diffusion UNet for denoising. Finally, the refined image $\hat{I}$ is decoded from the denoised latent $\hat{w}$.
Additionally, to account for per-view quality variations, we introduce a pre-defined view-dependent weight $f(p)$ to the reconstruction loss.
Empirically, we select a sine scheduler for pose-dependent weight, formulated as $f(p) = \sin (\pi \cdot \text{d}(x_p, x_0))$, where $x_0$ is the camera center of the first frame and $\text{d}(\cdot, \cdot)$ is the normalized L2 distance.
In total, the diffusion refinement loss $\mathcal{L}_\text{ref}$ is formulated as
\begin{equation}
    \mathcal{L}_{\text{ref}} = 
    f(p)\cdot(\mathcal{L}_\text{L1}(I_{t,p}, \hat I) + \lambda \cdot \mathcal{L}_\text{LPIPS} (I_{t,p}, \hat I)),
\end{equation}
where $\mathcal{L}_\text{LPIPS}(\cdot, \cdot)$ is the perceptual loss and $\mathcal{L}_\text{L1}(\cdot, \cdot)$ is the pixel-wise L$_1$ loss. 
To preserve the coarse geometry and texture from Stage II, we use this loss to fine-tune the coarse deformation field while keeping canonical Gaussians frozen.
To avoid error accumulation and unstable supervision, we conduct one-pass refinement: for each view/timestamp, the rendered image at the first iteration are used as the input of diffusion, and the refinement output is shared with all the iterations later.

\section{Experiments}
\begin{figure}[p]
    \centering
    \vspace{-3mm}
    \includegraphics[width=1.0\linewidth]{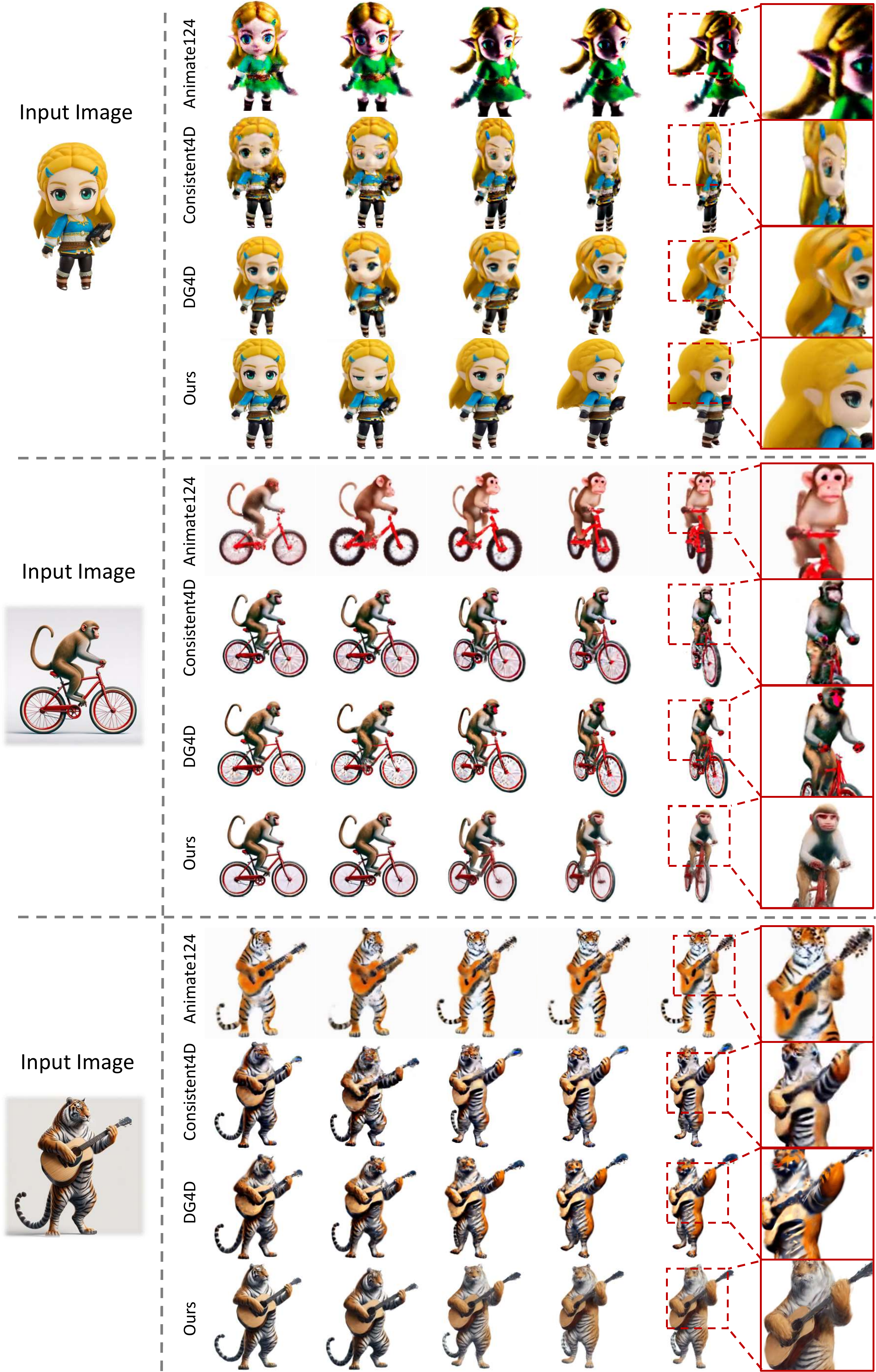}
    \caption{
    Comparison with Animate124~\cite{zhao2023animate124}, Consistent4D~\cite{jiang2023consistent4d}, and DreamGaussian4D (DG4D)~\cite{ren2023dreamgaussian4d} in three cases \texttt{zelda}, \texttt{monkey-bike} and \texttt{tiger-guitar} (better zoom in). The first two columns show the animation results in the same view, and the 3-5 columns demonstrate three other views. The last column illustrates the zoom-in image of the last rendered view.
    }
    \label{fig:comp}
\end{figure}

\subsection{Experimental Settings}
\label{subsec: exp_set}

\paragraph{Implementation details.}
In Stage I, we use SVD-img2vid-xl~\cite{blattmann2023stable} to generate 25-frame videos. For multi-view generation, we employ SV3D$^p$ conditioned on a camera pose sequence, \ie, 21 azimuth angles (360$^\circ$ evenly divided) and a fixed 0$^\circ$ elevation. All images are set to a resolution of 576$\times$576. In Stage III, we use SDXL-Turbo~\cite{sauer2023adversarial} with small strength (0.167) to provide the diffusion prior. For more reproduction details, please refer to the optimization settings in \cref{app:opt}.

\begin{figure}[t]
    \centering
    \vspace{-2mm}
    \includegraphics[width=1.0\linewidth]{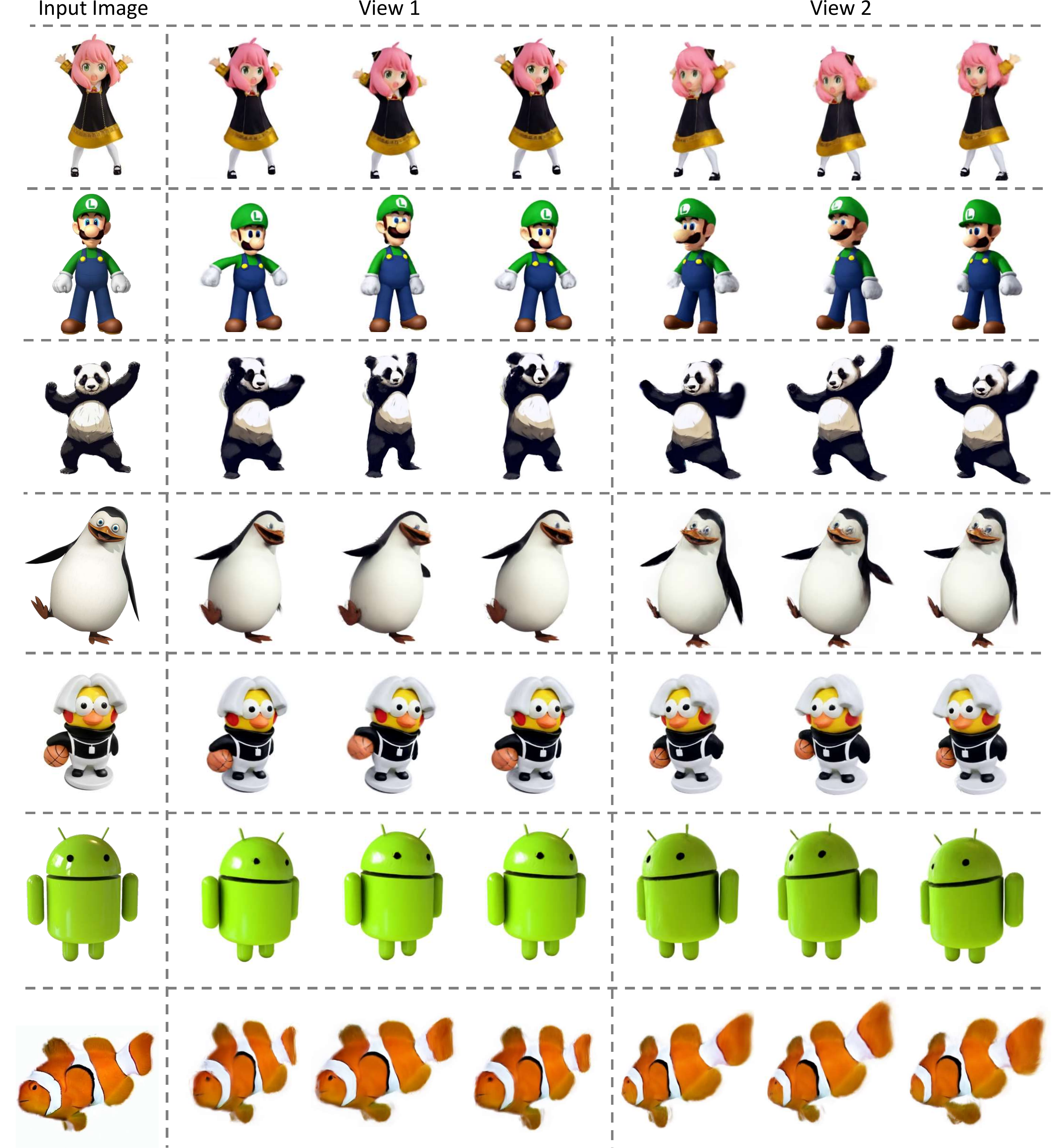}
    \caption{\textbf{Qualitative results of our generated 4D objects.} We present three consecutive frames rendered from our 4D model from two different views.}
    \label{fig:res}
    \vspace{-3mm}
\end{figure}

\paragraph{Evaluation metrics.}
Following previous methods~\cite{ren2023dreamgaussian4d, zhao2023animate124}, we use CLIP-I score that measures the cosine similarity of CLIP~\cite{radford2021learning} embedding of the given image and the rendered views.
We also conduct a user preference study to evaluate the 3D appearance, image-3D alignment, motion realism, motion range, and overall 4D quality. More details on user study can be found in \cref{app:user}.

\paragraph{Baselines.}
We compare our results with the state-of-the-art image-to-4D methods: Animate124~\cite{zhao2023animate124} and DreamGaussian4D~\cite{ren2023dreamgaussian4d}.
We also compare with the state-of-the-art video-to-4D method Consistent4D~\cite{jiang2023consistent4d}.
For a fair comparison, we feed the SVD-generated videos (exactly the same as ours) to Consistent4D for direct video-to-4D generation.

\subsection{Results}
\newpage

\paragraph{Qualitative results.}
\cref{fig:comp} demonstrates three cases for comparison between our \Ours{} and the baselines~\cite{ren2023dreamgaussian4d, zhao2023animate124, jiang2023consistent4d}.
Our generated results present better image-4D alignment and more realistic 3D appearance, especially in facial details.
Animate124 can not generate image-aligned 4D models because of its strong text guidance.
Consistent4D and DreamGaussian4D produce models with over-saturated and non-realistic appearance (especially in face) due to the inherent limitation of score distillation algorithm.
\cref{fig:res} shows detailed results produced by \Ours{}.
For each case, we present three temporal-continuous rendered frames from two views.
Illustration of more cases can be found in \cref{app:res}.
Rendered videos are provided in the supplementary for better motion visualization.

\begin{table}[t]
    \centering
    \caption{\textbf{User study on image-to-4D methods.} Each number represents the percentage of user preference. Error bars correspond to the 95.6$\%$ confidence interval.
    \textbf{Bold} denotes the best result.}
    \resizebox{\linewidth}{!}{
    \begin{tabular}{l|ccccc}
        \toprule
        Method &  Overall Quality & Ref. View Alignment & 3D Appearance &  Motion Realism & Motion Range \\
        \midrule
        Animate124~\cite{zhao2023animate124} 
        & 1.10 \small{$\pm$1.24} 
 & 2.24 \small{$\pm$1.77}
 &  1.65 \small{$\pm$1.52}
 & 2.19 \small{$\pm$1.75}
 & 5.39 \small{$\pm$2.70} \\
    Consistent4D~\cite{jiang2023consistent4d} & 
        3.88 \small{$\pm$2.31}
        & 5.00 \small{$\pm$2.60}
        & 3.99 \small{$\pm$2.34} 
        & 5.66 \small{$\pm$2.76} 
        & 8.67 \small{$\pm$3.36} \\
        DreamGaussian4D~\cite{ren2023dreamgaussian4d} & 11.27 \small{$\pm$3.78} &  12.17 \small{$\pm$2.91} & 10.29 \small{$\pm$3.63} & 
        15.42 \small{$\pm$4.32} & 
        39.96 \small{$\pm$5.83} \\
        \Ours{} (Ours) & \textbf{83.75} \small{$\pm$4.41} & \textbf{80.59} \small{$\pm$4.73} & \textbf{84.07} \small{$\pm$4.37} & \textbf{76.73} \small{$\pm$5.05} & \textbf{45.98} \small{$\pm$5.93} \\
        \bottomrule
    \end{tabular}
    }
    \label{tab:user}
\end{table}

\setlength{\intextsep}{0pt}
\begin{wraptable}{r}{0.3\textwidth}
\caption{\textbf{Quantitative results.}
}
\fontsize{8}{9}\selectfont
\begin{center}
\vspace{-3mm}
\resizebox{\linewidth}{!}{
\begin{tabular}{lc}
\toprule

Method & CLIP-I~$\uparrow$ \\  
\midrule 
Animate124~\cite{zhao2023animate124} & 0.8544 \\
Consistent4D~\cite{jiang2023consistent4d} & {0.9214} \\
DreamGaussian4D~\cite{ren2023dreamgaussian4d} & {0.9227} \\
\midrule
\Ours{} (Ours) & \textbf{0.9535} \\

\bottomrule
\label{tab:clip}
\end{tabular}
}
\end{center}
\end{wraptable}

\paragraph{Quantitative results \& User study.}
\cref{tab:clip} shows that our method has the highest CLIP-I score, which means the rendered images are more semantically similar to the reference image.
User study (\cref{tab:user}) shows that the recipients are overwhelmingly inclined towards the 4D results generated by our framework.
Almost 80\% of the participants think our method is superior in overall quality, reference view consistency, 3D appearance, and motion realism. 
Meanwhile, our motion range is on par with the strongest baseline, which is further discussed in \cref{sec:conclusion}. 

\subsection{Ablation Studies}

\begin{figure}[t]
    \centering
    \includegraphics[width=1.0\linewidth]{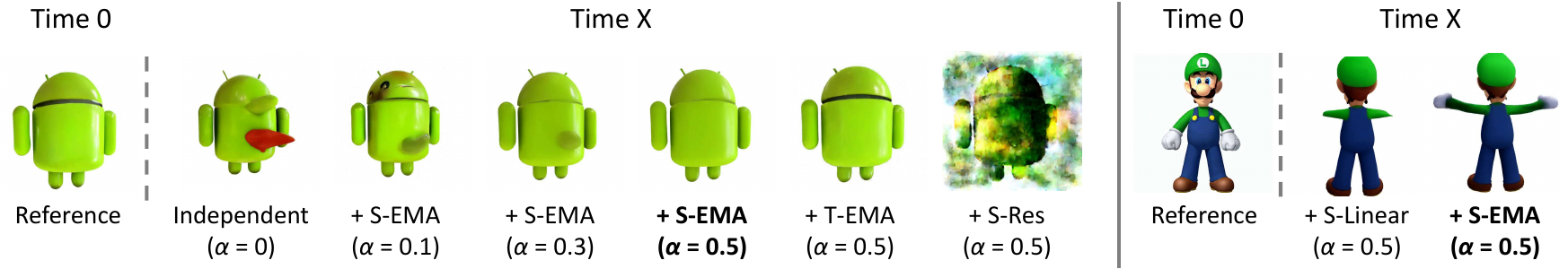}
    \caption{\textbf{Ablation on attention injection}. Video generation results are shown with two cases at time 0 and a timestamp X afterward. ``S-'' and ``T-'' stand for operations in spatial and temporal attention layers of SV3D, respectively. ``EMA'' denotes the proposed KV latent blending with Exponential Moving Average. ``Linear'' denotes KV blending with only the first frame. ``Res'' denotes injection on residual connection instead of KV. $\alpha$ is the blending weight. Different degrees of temporal inconsistency can be observed in all settings except ours (S-EMA, $\alpha=0.5$).}
    \label{fig:ablation_attn}
\end{figure}
\setlength{\intextsep}{0pt}
\begin{wrapfigure}{r}{0.4\textwidth}
  \centering
  \resizebox{\linewidth}{!}{
  \includegraphics{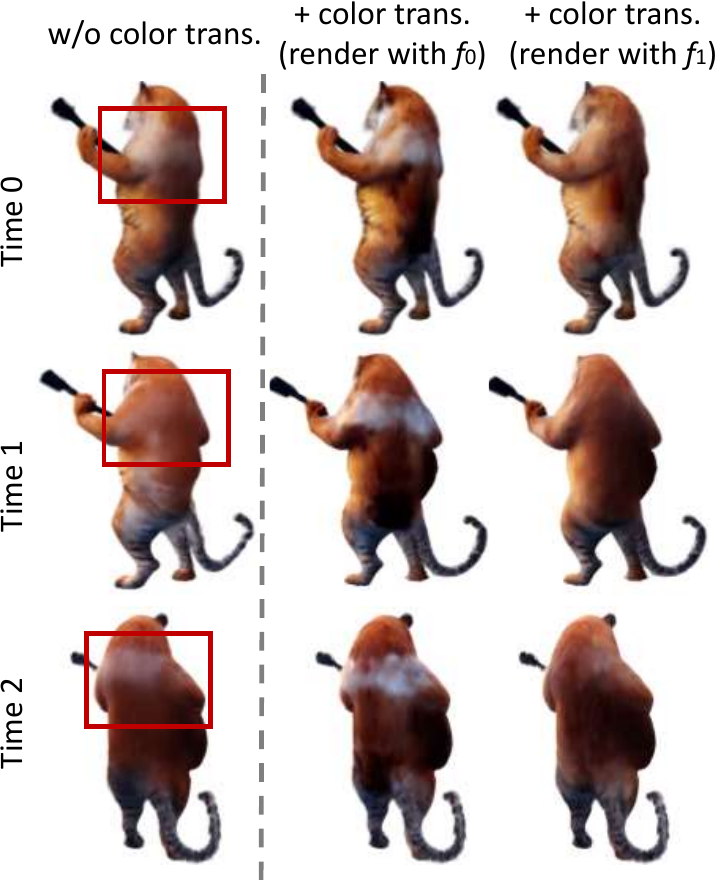}
  }
  \caption{\textbf{Effects of color transformation.} Our color affine transformation effectively disentangles the texture variation at different timestamps, enabling the rendering of color-consistent dynamics with arbitrary time-specific feature $\bm{f}_t$.}
  \vspace{-2mm}
  \label{fig:ablation_color}
\end{wrapfigure}

\paragraph{Attention injection.}
In \cref{fig:ablation_attn}, we explore the effect of attention injection by generating videos (Stage I) with different blending weight $\alpha$ and replacing our spatial KV latent blending with three variants:
\textbf{1)} \emph{T-EMA}: similar KV blending is adopted but in temporal attention layers of SV3D, \ie, one frame is blended with all views of the reference timestamp, which results in almost identical (static) results.
\textbf{2)} \emph{S-Res}: the residual term (skip connection) instead of KV latent is blended, which leads to collapse results.
\textbf{3)} \emph{S-Linear}: KV blending is used but only with the first frame. Without EMA, the diffusion model shows degraded generating capability for large motions departing from the reference frame (\texttt{luigi} in \cref{fig:ablation_attn} right).
Moreover, we observe that the temporal consistency is highly sensitive to blending weight $\alpha$.
For comparison, without any attention injection strategy ($\alpha=0$), views of different timestamps are generated independently, leading to dramatic temporal inconsistency in the back view of \texttt{android} (\cref{fig:ablation_attn} left). Our proposed spatial KV blending with EMA effectively improves the consistency when $\alpha$ is increased to $0.5$. Please refer to \cref{app:ablation} for the dynamic attenuation phenomenon when $\alpha>0.5$.

\paragraph{Color transformation.}
\cref{fig:ablation_color} shows the effectiveness of our proposed color transformation in Stage II. Dynamic 3DGS typically models all the inter-frame texture diversity as part of time-specific deformation. With color affine transformation, we manage to disentangle unwanted color inconsistencies and render temporal-consistent texture details from whichever timestamp we select.

\begin{figure}[t]
    \begin{subfigure}{0.35\textwidth}
        \centering
        \includegraphics[width=\linewidth]{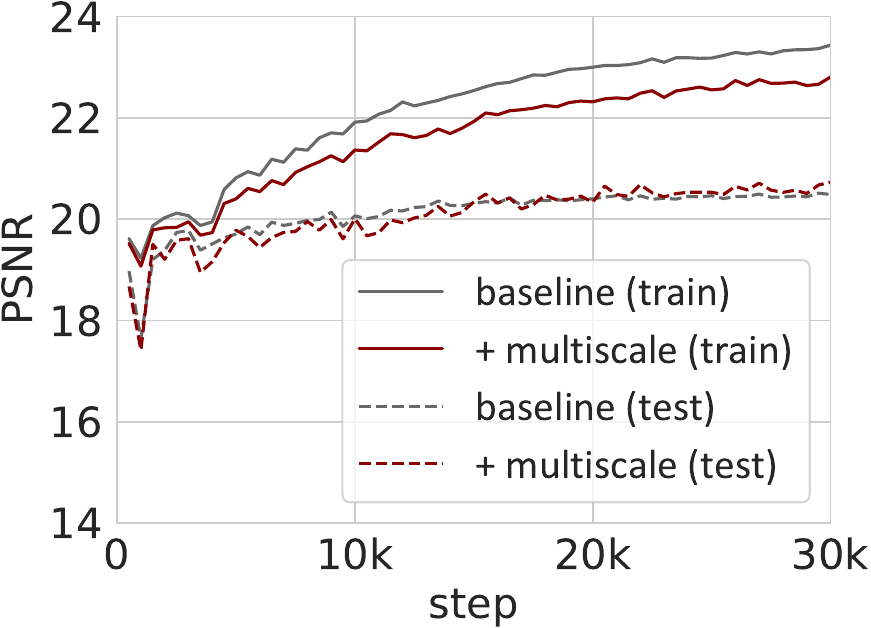}
        \caption{Reconstruction quality comparison.}
    \end{subfigure}
    \begin{subfigure}{0.64\textwidth}
        \centering
        \includegraphics[width=\linewidth]{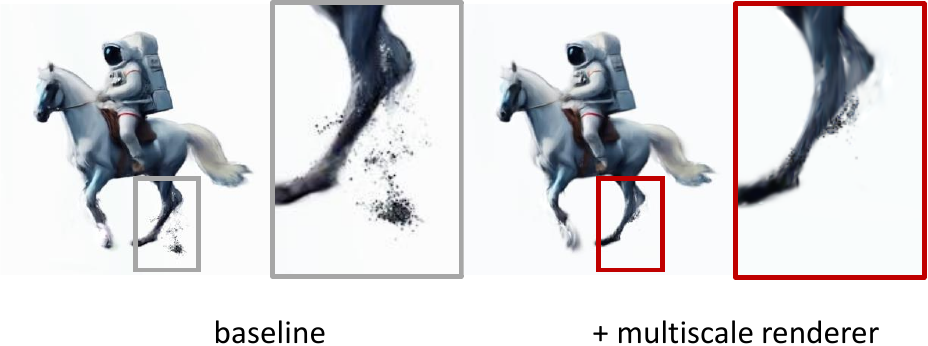}
        \caption{Visual details illustration.}
    \end{subfigure}
    \caption{\textbf{Effects of multiscale renderer.}
    \textbf{(a)} demonstrates the training (solid line) / test (dashed line) curve before (\textcolor{darkgray}{dark gray}) and after (\textcolor{darkred}{dark red}) adding the multiscale renderer.
    The multiscale rendering avoids the meaningless overfitting of our model (lower training PSNR, but comparative or even higher test PSNR). 
    \textbf{(b)} shows one viewpoint of rendering for case \texttt{astronaut-horse}. The multiscale render effectively prevents the model from overfitting to noise introduced in video diffusions.
    }
    \label{fig:ms}
\end{figure}

\paragraph{Multiscale renderer.}
\cref{fig:ms} shows the effectiveness of our multiscale renderer of Stage II.
We show the training and test PSNR during optimization in the left panel.
It can be observed that the multiscale renderer plays the role of a regularizer that effectively avoids model overfitting (lower training PSNR and similar test PSNR).
The qualitative result in the right panel illustrates that this design avoids overfitting to the noise introduced by the video diffusions.

\begin{figure}[t]
    \centering
    \includegraphics[width=1.0\linewidth]{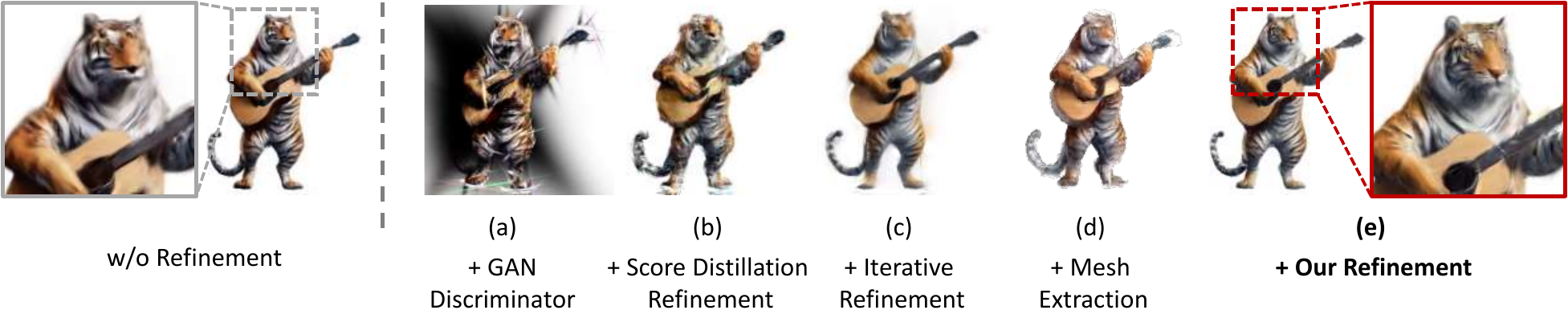}
    \caption{
    \textbf{Ablation on different refinement methods.} The leftmost column shows the image rendered by 4D model after the first two stage optimization for the case \texttt{tiger-guitar}.
    Panels (a) - (e) demonstrate different refinement methods aimed at addressing the semantic defects.
    However, only the one-pass refinement (ours) successfully adds facial details while keeping the original structure.
    }
    \label{fig:ablation_diffu}
    \vspace{-4mm}
\end{figure}

\paragraph{Refinement strategies.}
\cref{fig:ablation_diffu} illustrates ablations of refinement by comparing the visual details before and after applying various refinement techniques.
\textbf{(a)} \emph{Adversarial training}:
many previous works~\cite{wu2024rafe, chen2024it3d, roessle2023ganerf} leverage a GAN discriminator to optimize neural fields. However, we observe that although the discriminator loss converges quickly, the Gaussian points gradually diverge from the object surface, resulting in rendered images turning black after several iterations.
\textbf{(b)} \emph{SDS (score distillation refinement)}:
SDS seeks a single mode for text-aligned 4D representation, leading to unsuccessful refinement.
\textbf{(c)} \emph{Iterative refinement}: 
InstructN2N~\cite{haque2023instruct} iteratively updates the supervised dataset (each image is refined for multiple times, different from our \emph{one-pass refinement}) for 3D scene editing. In our task, the diverse outputs from diffusion model result in blurred 4D model under pixel-wise supervision.
\textbf{(d)} \emph{Textured mesh extraction}~\cite{tang2023dreamgaussian}: experiments show that meshes extracted from 3D Gaussians are not watertight and smooth~\cite{tang2023dreamgaussian, huang20242d}, leading to incoherent appearance.
\textbf{(e)} \emph{One-pass refinement} (Ours): in this way, the refined (supervised) images achieve a balance between detail restoration and preservation of structural integrity and consistency. It can be observed that reasonable details are introduced in regions with noise or semantic defects.

\section{Conclusion and Discussion}
\label{sec:conclusion}

\noindent\textbf{Conclusion.}
In this paper, we propose \Ours{}, a novel framework for 4D generation from a single image.
This approach departs from previous score-distillation-based methodologies, promising not only intrinsic immunity against problems like over-saturation but also capabilities for consistent visual details and dynamics.
We first equip the video diffusions with a training-free attention injection strategy to explicitly generate consistent dynamics and multi-views of the given object.
Then a coarse-to-fine 4D optimization scheme is introduced to further address practical challenges in synthesized videos.
Qualitative and quantitative results demonstrate that \Ours{} produces 4D objects with more realistic and higher-quality appearance and motion compared with the baselines.

\noindent\textbf{Limitations \& Future work.}
One limitation is that our framework can not generate high-dynamic motion due to the limited capability of the base image-to-video model~\cite{blattmann2023stable} and the consistency-motion trade-off in our attention injection strategy.
Another problem lies in the multi-view diffusion model~\cite{voleti2024sv3d}, which currently struggles to apply precise camera pose conditioning, leading to unsatisfactory reconstruction.
One solution for dynamics is to leverage more advanced video diffusions to generate high-quality and high-dynamic video frames.
Future work could also incorporate adaptive camera pose techniques~\cite{fu2023colmap, smith2024flowmap} in 4D reconstruction to further improve the robustness.

\noindent\textbf{Broader impact.}
Our work transforms a single image into a dynamic 3D object, which could present challenges related to copyright issues, as well as wider implications for privacy and data security.

\bibliographystyle{unsrt}
\bibliography{main, bibliography_long, bibliography_custom, egbib_nips2024}

\begin{thebibliography}{10}

\bibitem{Brown2020NeurIPS}
Tom~B. Brown, Benjamin Mann, Nick Ryder, Melanie Subbiah, Jared Kaplan, Prafulla Dhariwal, Arvind Neelakantan, Pranav Shyam, Girish Sastry, Amanda Askell, Sandhini Agarwal, Ariel Herbert{-}Voss, Gretchen Krueger, Tom Henighan, Rewon Child, Aditya Ramesh, Daniel~M. Ziegler, Jeffrey Wu, Clemens Winter, Christopher Hesse, Mark Chen, Eric Sigler, Mateusz Litwin, Scott Gray, Benjamin Chess, Jack Clark, Christopher Berner, Sam McCandlish, Alec Radford, Ilya Sutskever, and Dario Amodei.
\newblock Language models are few-shot learners.
\newblock In {\em NeurIPS}, 2020.

\bibitem{Touvron2023ARXIV}
Hugo Touvron, Thibaut Lavril, Gautier Izacard, Xavier Martinet, Marie{-}Anne Lachaux, Timoth{\'{e}}e Lacroix, Baptiste Rozi{\`{e}}re, Naman Goyal, Eric Hambro, Faisal Azhar, Aur{\'{e}}lien Rodriguez, Armand Joulin, Edouard Grave, and Guillaume Lample.
\newblock {Llama}: Open and efficient foundation language models.
\newblock {\em arXiv preprint arXiv:2302.13971}, 2023.

\bibitem{OpenATI2023ARXIV}
OpenAI.
\newblock {GPT}-4 technical report.
\newblock {\em arXiv preprint arXiv:2303.08774}, 2023.

\bibitem{Ramesh2021ICML}
Aditya Ramesh, Mikhail Pavlov, Gabriel Goh, Scott Gray, Chelsea Voss, Alec Radford, Mark Chen, and Ilya Sutskever.
\newblock Zero-shot text-to-image generation.
\newblock In {\em ICML}, 2021.

\bibitem{Rombach2022CVPR}
Robin Rombach, Andreas Blattmann, Dominik Lorenz, Patrick Esser, and Bj{\"{o}}rn Ommer.
\newblock High-resolution image synthesis with latent diffusion models.
\newblock In {\em CVPR}, 2022.

\bibitem{Sauer2023ICML}
Axel Sauer, Tero Karras, Samuli Laine, Andreas Geiger, and Timo Aila.
\newblock {StyleGAN-T}: Unlocking the power of {GANs} for fast large-scale text-to-image synthesis.
\newblock In {\em ICML}, 2023.

\bibitem{Skorokhodov2022CVPR}
Ivan Skorokhodov, Sergey Tulyakov, and Mohamed Elhoseiny.
\newblock {StyleGAN-V}: {A} continuous video generator with the price, image quality and perks of {StyleGAN2}.
\newblock In {\em CVPR}, 2022.

\bibitem{Bahmani2022ARXIV}
Sherwin Bahmani, Jeong~Joon Park, Despoina Paschalidou, Hao Tang, Gordon Wetzstein, Leonidas~J. Guibas, Luc~Van Gool, and Radu Timofte.
\newblock {3D-Aware} video generation.
\newblock {\em TMLR}, 2023.

\bibitem{Singer2022ARXIV}
Uriel Singer, Adam Polyak, Thomas Hayes, Xi~Yin, Jie An, Songyang Zhang, Qiyuan Hu, Harry Yang, Oron Ashual, Oran Gafni, Devi Parikh, Sonal Gupta, and Yaniv Taigman.
\newblock {Make-A-Video}: Text-to-video generation without text-video data.
\newblock {\em ICLR}, 2023.

\bibitem{chan2022efficient}
Eric~R Chan, Connor~Z Lin, Matthew~A Chan, Koki Nagano, Boxiao Pan, Shalini De~Mello, Orazio Gallo, Leonidas~J Guibas, Jonathan Tremblay, Sameh Khamis, et~al.
\newblock Efficient geometry-aware {3D} generative adversarial networks.
\newblock In {\em CVPR}, 2022.

\bibitem{poole2022dreamfusion}
Ben Poole, Ajay Jain, Jonathan~T Barron, and Ben Mildenhall.
\newblock {DreamFusion}: Text-to-{3D} using {2D} diffusion.
\newblock In {\em ICLR}, 2023.

\bibitem{Lin2023CVPR}
Chen{-}Hsuan Lin, Jun Gao, Luming Tang, Towaki Takikawa, Xiaohui Zeng, Xun Huang, Karsten Kreis, Sanja Fidler, Ming{-}Yu Liu, and Tsung{-}Yi Lin.
\newblock {Magic3D}: High-resolution text-to-{3D} content creation.
\newblock In {\em CVPR}, 2023.

\bibitem{tang2023dreamgaussian}
Jiaxiang Tang, Jiawei Ren, Hang Zhou, Ziwei Liu, and Gang Zeng.
\newblock {DreamGaussian}: Generative gaussian splatting for efficient {3D} content creation.
\newblock In {\em ICLR}, 2024.

\bibitem{ren2023dreamgaussian4d}
Jiawei Ren, Liang Pan, Jiaxiang Tang, Chi Zhang, Ang Cao, Gang Zeng, and Ziwei Liu.
\newblock {DreamGaussian4D}: Generative {4D} gaussian splatting.
\newblock {\em arXiv preprint arXiv:2312.17142}, 2023.

\bibitem{zhao2023animate124}
Yuyang Zhao, Zhiwen Yan, Enze Xie, Lanqing Hong, Zhenguo Li, and Gim~Hee Lee.
\newblock Animate124: Animating one image to {4D} dynamic scene.
\newblock {\em arXiv preprint arXiv:2311.14603}, 2023.

\bibitem{jiang2023consistent4d}
Yanqin Jiang, Li~Zhang, Jin Gao, Weimin Hu, and Yao Yao.
\newblock {Consistent4D}: Consistent 360\textdegree~dynamic object generation from monocular video.
\newblock In {\em ICLR}, 2024.

\bibitem{singer2023text}
Uriel Singer, Shelly Sheynin, Adam Polyak, Oron Ashual, Iurii Makarov, Filippos Kokkinos, Naman Goyal, Andrea Vedaldi, Devi Parikh, Justin Johnson, et~al.
\newblock Text-to-{4D} dynamic scene generation.
\newblock {\em arXiv preprint arXiv:2301.11280}, 2023.

\bibitem{Bailey2020ArtInAmerica}
Jason Bailey.
\newblock The tools of generative art, from flash to neural networks.
\newblock {\em Art in America}, 8, 2020.

\bibitem{yin20234dgen}
Yuyang Yin, Dejia Xu, Zhangyang Wang, Yao Zhao, and Yunchao Wei.
\newblock {4DGen}: Grounded {4D} content generation with spatial-temporal consistency.
\newblock {\em arXiv preprint arXiv:2312.17225}, 2023.

\bibitem{xu2024comp4d}
Dejia Xu, Hanwen Liang, Neel~P Bhatt, Hezhen Hu, Hanxue Liang, Konstantinos~N Plataniotis, and Zhangyang Wang.
\newblock {Comp4D}: {LLM}-guided compositional {4D} scene generation.
\newblock {\em arXiv preprint arXiv:2403.16993}, 2024.

\bibitem{bahmani2024tc4d}
Sherwin Bahmani, Xian Liu, Yifan Wang, Ivan Skorokhodov, Victor Rong, Ziwei Liu, Xihui Liu, Jeong~Joon Park, Sergey Tulyakov, Gordon Wetzstein, et~al.
\newblock {TC4D}: Trajectory-conditioned text-to-{4D} generation.
\newblock {\em arXiv preprint arXiv:2403.17920}, 2024.

\bibitem{bahmani20234d}
Sherwin Bahmani, Ivan Skorokhodov, Victor Rong, Gordon Wetzstein, Leonidas Guibas, Peter Wonka, Sergey Tulyakov, Jeong~Joon Park, Andrea Tagliasacchi, and David~B Lindell.
\newblock {4D}-fy: Text-to-{4D} generation using hybrid score distillation sampling.
\newblock In {\em CVPR}, 2024.

\bibitem{zheng2024unified}
Yufeng Zheng, Xueting Li, Koki Nagano, Sifei Liu, Otmar Hilliges, and Shalini~De Mello.
\newblock A unified approach for text- and image-guided {4D} scene generation.
\newblock In {\em CVPR}, 2024.

\bibitem{ling2023align}
Huan Ling, Seung~Wook Kim, Antonio Torralba, Sanja Fidler, and Karsten Kreis.
\newblock Align your {G}aussians: Text-to-4{D} with dynamic 3{D} gaussians and composed diffusion models.
\newblock In {\em CVPR}, 2024.

\bibitem{wang2023prolificdreamer}
Zhengyi Wang, Cheng Lu, Yikai Wang, Fan Bao, Chongxuan Li, Hang Su, and Jun Zhu.
\newblock {ProlificDreamer}: High-fidelity and diverse text-to-{3D} generation with variational score distillation.
\newblock {\em NeurIPS}, 2023.

\bibitem{armandpour2023re}
Mohammadreza Armandpour, Ali Sadeghian, Huangjie Zheng, Amir Sadeghian, and Mingyuan Zhou.
\newblock Re-imagine the negative prompt algorithm: Transform {2D} diffusion into {3D}, alleviate janus problem and beyond.
\newblock {\em arXiv preprint arXiv:2304.04968}, 2023.

\bibitem{voleti2024sv3d}
Vikram Voleti, Chun-Han Yao, Mark Boss, Adam Letts, David Pankratz, Dmitry Tochilkin, Christian Laforte, Robin Rombach, and Varun Jampani.
\newblock {SV3D}: Novel multi-view synthesis and {3D} generation from a single image using latent video diffusion.
\newblock {\em arXiv preprint arXiv:2403.12008}, 2024.

\bibitem{blattmann2023stable}
Andreas Blattmann, Tim Dockhorn, Sumith Kulal, Daniel Mendelevitch, Maciej Kilian, Dominik Lorenz, Yam Levi, Zion English, Vikram Voleti, Adam Letts, et~al.
\newblock Stable video diffusion: Scaling latent video diffusion models to large datasets.
\newblock {\em arXiv preprint arXiv:2311.15127}, 2023.

\bibitem{brooks2024video}
T~Brooks, B~Peebles, C~Homes, W~DePue, Y~Guo, L~Jing, D~Schnurr, J~Taylor, T~Luhman, E~Luhman, et~al.
\newblock Video generation models as world simulators.
\newblock {\em OpenAI Blog}, 2024.

\bibitem{4dgs}
Guanjun Wu, Taoran Yi, Jiemin Fang, Lingxi Xie, Xiaopeng Zhang, Wei Wei, Wenyu Liu, Qi~Tian, and Xinggang Wang.
\newblock {{4D}} gaussian splatting for real-time dynamic scene rendering.
\newblock In {\em CVPR}, 2024.

\bibitem{yang2023deformable}
Ziyi Yang, Xinyu Gao, Wen Zhou, Shaohui Jiao, Yuqing Zhang, and Xiaogang Jin.
\newblock Deformable {3D} gaussians for high-fidelity monocular dynamic scene reconstruction.
\newblock {\em arXiv preprint arXiv:2309.13101}, 2023.

\bibitem{huang2023sc}
Yi-Hua Huang, Yang-Tian Sun, Ziyi Yang, Xiaoyang Lyu, Yan-Pei Cao, and Xiaojuan Qi.
\newblock {SC-GS}: Sparse-controlled gaussian splatting for editable dynamic scenes.
\newblock In {\em CVPR}, 2024.

\bibitem{hoDenoisingDiffusionProbabilistic2020}
Jonathan Ho, Ajay Jain, and Pieter Abbeel.
\newblock Denoising diffusion probabilistic models.
\newblock In {\em NeurIPS}, 2020.

\bibitem{hoVideoDiffusionModels2022}
Jonathan Ho, Tim Salimans, Alexey Gritsenko, William Chan, Mohammad Norouzi, and David~J. Fleet.
\newblock Video diffusion models.
\newblock {\em arXiv preprint arXiv:2204.03458}, 2022.

\bibitem{singerMakeAVideoTexttoVideoGeneration2022}
Uriel Singer, Adam Polyak, Thomas Hayes, Xi~Yin, Jie An, Songyang Zhang, Qiyuan Hu, Harry Yang, Oron Ashual, Oran Gafni, et~al.
\newblock Make-a-video: Text-to-video generation without text-video data.
\newblock {\em arXiv preprint arXiv:2209.14792}, 2022.

\bibitem{hoImagenVideoHigh2022}
Jonathan Ho, William Chan, Chitwan Saharia, Jay Whang, Ruiqi Gao, Alexey Gritsenko, Diederik~P. Kingma, Ben Poole, Mohammad Norouzi, and David~J. Fleet.
\newblock {Video}: {{High}} definition video generation with diffusion models.
\newblock {\em arXiv preprint arXiv:2210.02303}, 2022.

\bibitem{yinNUWAXLDiffusionDiffusion2023}
Shengming Yin, Chenfei Wu, Huan Yang, Jianfeng Wang, Xiaodong Wang, Minheng Ni, Zhengyuan Yang, Linjie Li, Shuguang Liu, and Fan Yang.
\newblock {{NUWA-XL}}: {{Diffusion}} over {{Diffusion}} for {{eXtremely Long Video Generation}}.
\newblock In {\em ACL}, 2023.

\bibitem{khachatryanText2VideoZeroTexttoImageDiffusion2023}
Levon Khachatryan, Andranik Movsisyan, Vahram Tadevosyan, Roberto Henschel, Zhangyang Wang, Shant Navasardyan, and Humphrey Shi.
\newblock {{Text2Video-Zero}}: Text-to-image diffusion models are zero-shot video generators.
\newblock In {\em ICCV}, 2023.

\bibitem{rombachHighResolutionImageSynthesis2022}
Robin Rombach, Andreas Blattmann, Dominik Lorenz, Patrick Esser, and Bj{\"o}rn Ommer.
\newblock High-resolution image synthesis with latent diffusion models.
\newblock In {\em CVPR}, 2022.

\bibitem{blattmannAlignYourLatents2023}
Andreas Blattmann, Robin Rombach, Huan Ling, Tim Dockhorn, Seung~Wook Kim, Sanja Fidler, and Karsten Kreis.
\newblock Align {{Your Latents}}: {{High-Resolution Video Synthesis With Latent Diffusion Models}}.
\newblock In {\em CVPR}, 2023.

\bibitem{guoAnimateDiffAnimateYour2023c}
Yuwei Guo, Ceyuan Yang, Anyi Rao, Yaohui Wang, Yu~Qiao, Dahua Lin, and Bo~Dai.
\newblock {{AnimateDiff}}: Animate your personalized text-to-image diffusion models without specific tuning.
\newblock In {\em ICLR}, 2024.

\bibitem{wan2023cad}
Ziyu Wan, Despoina Paschalidou, Ian Huang, Hongyu Liu, Bokui Shen, Xiaoyu Xiang, Jing Liao, and Leonidas Guibas.
\newblock {CAD}: Photorealistic {3D} generation via adversarial distillation.
\newblock In {\em CVPR}, 2024.

\bibitem{cai2023generative}
Shengqu Cai, Duygu Ceylan, Matheus Gadelha, Chun-Hao~Paul Huang, Tuanfeng~Yang Wang, and Gordon Wetzstein.
\newblock Generative rendering: Controllable {4D}-guided video generation with {2D} diffusion models.
\newblock {\em arXiv preprint arXiv:2312.01409}, 2023.

\bibitem{fridovich2023k}
Sara Fridovich-Keil, Giacomo Meanti, Frederik~Rahb{\ae}k Warburg, Benjamin Recht, and Angjoo Kanazawa.
\newblock {K-Planes}: Explicit radiance fields in space, time, and appearance.
\newblock In {\em CVPR}, 2023.

\bibitem{cao2023hexplane}
Ang Cao and Justin Johnson.
\newblock {HexPlane}: A fast representation for dynamic scenes.
\newblock In {\em CVPR}, 2023.

\bibitem{singer2022make}
Uriel Singer, Adam Polyak, Thomas Hayes, Xi~Yin, Jie An, Songyang Zhang, Qiyuan Hu, Harry Yang, Oron Ashual, Oran Gafni, et~al.
\newblock {Make-a-Video}: Text-to-video generation without text-video data.
\newblock {\em arXiv preprint arXiv:2209.14792}, 2022.

\bibitem{liu2023zero}
Ruoshi Liu, Rundi Wu, Basile Van~Hoorick, Pavel Tokmakov, Sergey Zakharov, and Carl Vondrick.
\newblock Zero-1-to-3: Zero-shot one image to {3D} object.
\newblock {\em arXiv preprint arXiv:2303.11328}, 2023.

\bibitem{liu2023one}
Minghua Liu, Chao Xu, Haian Jin, Linghao Chen, Zexiang Xu, Hao Su, et~al.
\newblock One-2-3-45: Any single image to 3d mesh in 45 seconds without per-shape optimization.
\newblock {\em arXiv preprint arXiv:2306.16928}, 2023.

\bibitem{shi2023MVDream}
Yichun Shi, Peng Wang, Jianglong Ye, Long Mai, Kejie Li, and Xiao Yang.
\newblock {MVDream}: Multi-view diffusion for {3D} generation.
\newblock In {\em ICLR}, 2024.

\bibitem{zhang2023adding}
Lvmin Zhang, Anyi Rao, and Maneesh Agrawala.
\newblock Adding conditional control to text-to-image diffusion models.
\newblock In {\em ICCV}, 2023.

\bibitem{3dgs}
Bernhard Kerbl, Georgios Kopanas, Thomas Leimkuehler, and George Drettakis.
\newblock {{3D}} gaussian splatting for real-time radiance field rendering.
\newblock {\em ACM TOG}, 2023.

\bibitem{volumnrender}
N.~Max.
\newblock Optical models for direct volume rendering.
\newblock {\em IEEE TVCG}, 1995.

\bibitem{guo2024motion}
Zhiyang Guo, Wengang Zhou, Li~Li, Min Wang, and Houqiang Li.
\newblock Motion-aware {3D} gaussian splatting for efficient dynamic scene reconstruction.
\newblock {\em arXiv preprint arXiv:2403.11447}, 2024.

\bibitem{li2022tava}
Ruilong Li, Julian Tanke, Minh Vo, Michael Zollh{\"o}fer, J{\"u}rgen Gall, Angjoo Kanazawa, and Christoph Lassner.
\newblock {TAVA}: Template-free animatable volumetric actors.
\newblock In {\em ECCV}, 2022.

\bibitem{darmon2024robust}
Fran{\c{c}}ois Darmon, Lorenzo Porzi, Samuel Rota-Bul{\`o}, and Peter Kontschieder.
\newblock Robust gaussian splatting.
\newblock {\em arXiv preprint arXiv:2404.04211}, 2024.

\bibitem{sauer2023adversarial}
Axel Sauer, Dominik Lorenz, Andreas Blattmann, and Robin Rombach.
\newblock Adversarial diffusion distillation.
\newblock {\em arXiv preprint arXiv:2311.17042}, 2023.

\bibitem{radford2021learning}
Alec Radford, Jong~Wook Kim, Chris Hallacy, Aditya Ramesh, Gabriel Goh, Sandhini Agarwal, Girish Sastry, Amanda Askell, Pamela Mishkin, Jack Clark, et~al.
\newblock Learning transferable visual models from natural language supervision.
\newblock In {\em ICML}, 2021.

\bibitem{wu2024rafe}
Zhongkai Wu, Ziyu Wan, Jing Zhang, Jing Liao, and Dong Xu.
\newblock {RaFE}: Generative radiance fields restoration.
\newblock {\em arXiv preprint arXiv:2404.03654}, 2024.

\bibitem{chen2024it3d}
Yiwen Chen, Chi Zhang, Xiaofeng Yang, Zhongang Cai, Gang Yu, Lei Yang, and Guosheng Lin.
\newblock {IT3D}: Improved text-to-{3D} generation with explicit view synthesis.
\newblock In {\em AAAI}, 2024.

\bibitem{roessle2023ganerf}
Barbara Roessle, Norman M{\"u}ller, Lorenzo Porzi, Samuel~Rota Bul{\`o}, Peter Kontschieder, and Matthias Nie{\ss}ner.
\newblock {GANeRF}: Leveraging discriminators to optimize neural radiance fields.
\newblock {\em ACM TOG}, 2023.

\bibitem{haque2023instruct}
Ayaan Haque, Matthew Tancik, Alexei~A Efros, Aleksander Holynski, and Angjoo Kanazawa.
\newblock Instruct-{NeRF2NeRF}: Editing {3D} scenes with instructions.
\newblock In {\em ICCV}, 2023.

\bibitem{huang20242d}
Binbin Huang, Zehao Yu, Anpei Chen, Andreas Geiger, and Shenghua Gao.
\newblock {2D} gaussian splatting for geometrically accurate radiance fields.
\newblock {\em ACM TOG}, 2024.

\bibitem{fu2023colmap}
Yang Fu, Sifei Liu, Amey Kulkarni, Jan Kautz, Alexei~A Efros, and Xiaolong Wang.
\newblock {COLMAP}-free {3D} gaussian splatting.
\newblock In {\em CVPR}, 2024.

\bibitem{smith2024flowmap}
Cameron Smith, David Charatan, Ayush Tewari, and Vincent Sitzmann.
\newblock {FlowMap}: High-quality camera poses, intrinsics, and depth via gradient descent.
\newblock {\em arXiv preprint arXiv:2404.15259}, 2024.

\bibitem{wu2023reconfusion}
Rundi Wu, Ben Mildenhall, Philipp Henzler, Keunhong Park, Ruiqi Gao, Daniel Watson, Pratul~P Srinivasan, Dor Verbin, Jonathan~T Barron, Ben Poole, et~al.
\newblock {ReconFusion}: {3D} reconstruction with diffusion priors.
\newblock In {\em CVPR}, 2024.

\bibitem{igs2gs}
Cyrus Vachha and Ayaan Haque.
\newblock Instruct-{GS2GS}: Editing {3D} gaussian splats with instructions, 2024.

\bibitem{zhou2023sparsefusion}
Zhizhuo Zhou and Shubham Tulsiani.
\newblock {SparseFusion}: Distilling view-conditioned diffusion for {3D} reconstruction.
\newblock In {\em CVPR}, 2023.

\bibitem{goodfellow2014gan}
Ian Goodfellow, Jean Pouget-Abadie, Mehdi Mirza, Bing Xu, David Warde-Farley, Sherjil Ozair, Aaron Courville, and Yoshua Bengio.
\newblock Generative adversarial nets.
\newblock In {\em NeurIPS}, 2014.

\bibitem{brooks2023instructpix2pix}
Tim Brooks, Aleksander Holynski, and Alexei~A Efros.
\newblock {InstructPix2Pix}: Learning to follow image editing instructions.
\newblock In {\em CVPR}, 2023.

\bibitem{podell2023sdxl}
Dustin Podell, Zion English, Kyle Lacey, Andreas Blattmann, Tim Dockhorn, Jonas M{\"u}ller, Joe Penna, and Robin Rombach.
\newblock {SDXL}: Improving latent diffusion models for high-resolution image synthesis.
\newblock In {\em ICLR}, 2024.

\end{thebibliography}
\clearpage

\clearpage

\appendix

{\LARGE \textbf{Appendix}}

\hspace{1pt}

In this appendix, we first present additional related works on 3D refinement (\cref{app:related}).
Then we provide detailed network specifications (\cref{app:network}).
Next, To ensure reproducibility and facilitate fair perceptual studies, we describe the experimental settings in detail (\cref{app:exp_set}).
Finally, we include extended ablation studies (\cref{app:ablation}) and additional visual results (\cref{app:res}) to demonstrate the robustness and superiority of our methods across various settings.

\section{More Related Works}
\label{app:related}

\paragraph{3D refinement with generative priors.}
To deal with view-inconsistency and low quality problems, many works~\cite{wu2024rafe, roessle2023ganerf, chen2024it3d, wu2023reconfusion, haque2023instruct, igs2gs, zhou2023sparsefusion} take advantages from generative priors, \eg, adversarial training~\cite{goodfellow2014gan} and score distillation sampling (SDS)~\cite{poole2022dreamfusion} to optimize the 3D representation.
GANeRF~\cite{roessle2023ganerf} refines the rendered images with an image-conditional generator and leverages the re-rendered image constraints to guide the NeRF optimization in the adversarial formulation.
InstructNeRF2NeRF~\cite{haque2023instruct} uses the text-conditioned image generator, InstructPix2pix~\cite{brooks2023instructpix2pix}, to edit the image rendered by pre-trained NeRF in an iterative manner and updates the underlying 3D representation with the edited images.
ReconFusion~\cite{wu2023reconfusion} uses the diffusion priors, Zero-123~\cite{liu2023zero}, as a drop-in regularizer to enhance the 3D reconstruction performance, especially for sparse-view scenarios.
In contrast to directly optimizing the implicit representation, another line of researches~\cite{tang2023dreamgaussian, ren2023dreamgaussian4d} first extracts the explicit textured mesh, and then refine the texture in UV-space with diffusion prior and differentiable rendering.
In our paper, in consideration of the artifacts generated in video diffusion, we extend the refinement techniques to the 4D representation.

\section{Network Details}
\label{app:network}

In this section, we unpack the network design in \cref{fig:framework}.

\begin{figure}[t]
    \centering
    \includegraphics[width=1.0\linewidth]{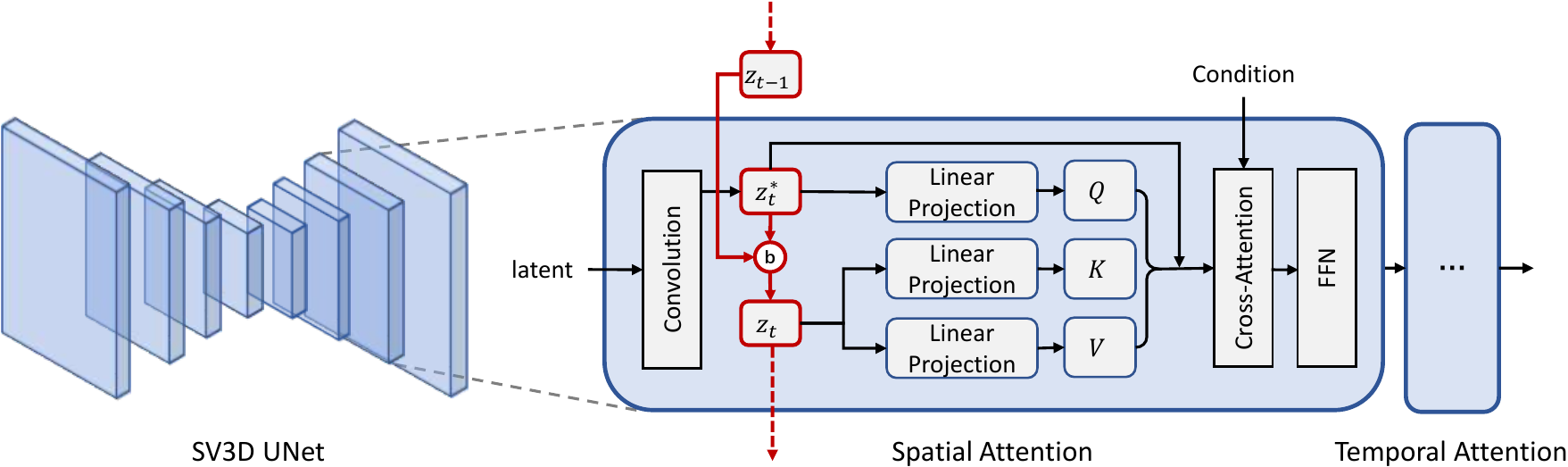}
    \caption{\textbf{Network details of Attention Injection.}  \circled{b} denotes the EMA blending operator mentioned in \cref{subsec:videogen}.
    $z_t^*$ is the multi-view latent at current timestamp $t$, and $z_t$ is the blended latent.
    Previous visual information is injected into the current latent by modifying the original spatial self-attention mechanism.
    }
    \label{fig:network_attn}
\end{figure}

\paragraph{Attention injection.}
In \cref{subsec:videogen}, we exploit the attention injection strategy to alleviate the temporal difference between multi-view diffusion models.
\cref{fig:network_attn} illustrates its network details: in each spatial attention layer, we replace the self-attention by simultaneously considering the current $z^*_t$ and previous visual information with EMA.

\begin{figure}[t]
    \centering
    \includegraphics[width=1.0\linewidth]{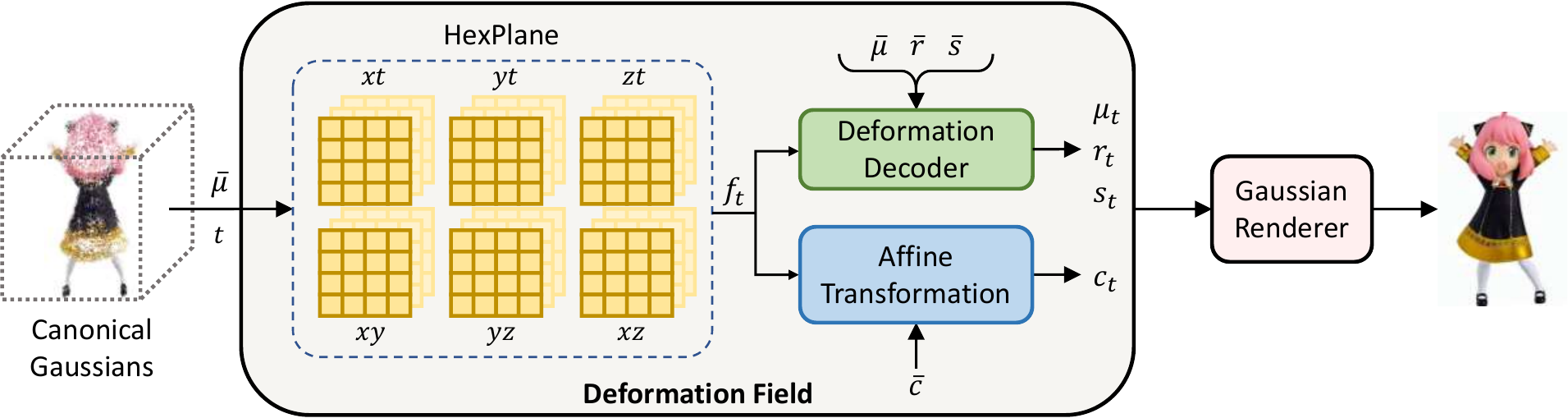}
    \caption{\textbf{Network architecture of our 4D representation.} 
    $\bar \mu, \bar r, \bar s$ represents the canonical Gaussian properties: 3D location, rotation and scale from the coarse stage training in 4DGS~\cite{4dgs}.
    The time-specific local feature $f_t$ is queried from the HexPlane~\cite{cao2023hexplane}, where the subscribe $t$ means the time-specific property.
    Different from vanilla 4DGS, we employ additional color affine transformation to obtain the time-specific color $c_t$.
    The geometric deformations are predicted by a lightweight decoder.
    Finally, the time-specific Gaussians are rendered to produce an image (right).
    }
    \label{fig:4dgs}
\end{figure}

\paragraph{Deformation field with color transformation.}
In \cref{subsec:recon}, we use color affine transformation to model the temporal texture variation.
\cref{fig:4dgs} shows the detailed architecture of it. 
We first query the time-specific feature $f_t$ from the HexPlane~\cite{cao2023hexplane} with the canonical Gaussian positions.
After that, the geometric deformations of Gaussian properties ($\mu$ location, $r$ rotation, and $s$ scale) are predicted with a lightweight decoder.
Additionally, we use the affine color transformation to model the temporal texture variations.
Finally, these deformed Gaussians are rendered into an image.

\section{Additional Experimental Settings}
\label{app:exp_set}

\subsection{Optimization Details}
\label{app:opt}

We report the optimization of 4D Gaussian splatting for the purpose of reproduction.
Basically, we follow the training recipe from 4DGS~\cite{4dgs} in the coarse 4D reconstruction stage.
In the semantic refinement stage (Stage III), we fine-tune 4DGS for 5k steps with Adam optimizer. The initial learning rate is set to 1e-4 with exponential decay.
The weight $\lambda$ in diffusion refinement loss is set to 0.5.
Our implementation is primarily based on the PyTorch framework and tested on a single NVIDIA RTX 3090 GPU.

\subsection{Reproduction, Data and Code}

We reproduced our baselines (Animate124~\cite{zhao2023animate124}, DreamGaussian4D~\cite{ren2023dreamgaussian4d}, and Consistent4D~\cite{jiang2023consistent4d}) using their official code. Additionally, we have included the input images and videos generated by SVD in the \emph{supplementary materials}. 
Apart from the data provided by Animate124 and DreamGaussian4D, we have added three more examples: \texttt{android}, \texttt{chicken-basketball}, and \texttt{penguin}. 
The code is also available in the \emph{supplementary materials}.

\subsection{User Study Details}
\label{app:user}

\begin{figure}[t]
    \centering
    \includegraphics[width=1.0\linewidth]{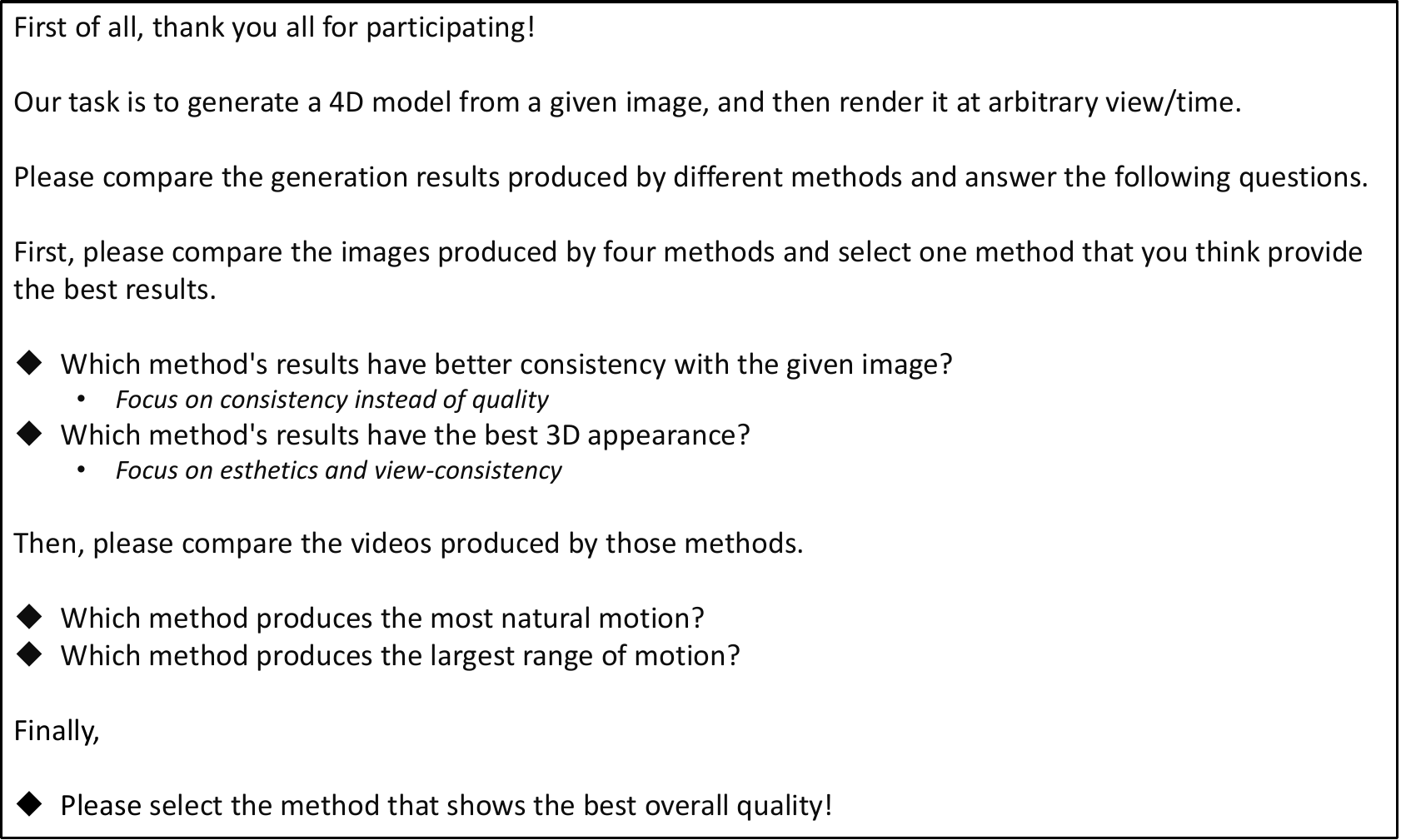}
    \caption{\textbf{Screenshot of our user study guidelines.} Each participant is asked to evaluate the images and videos rendered by 4 different methods with 5 metrics, \ie, reference view consistency, 3D appearance, motion realism, motion range, and overall quality.}
    \label{fig:user_study_detail}
\end{figure}

\begin{figure}[t]
    \centering
    \begin{subfigure}[b]{0.51\textwidth}
        \centering
        \includegraphics[width=1.0\textwidth]{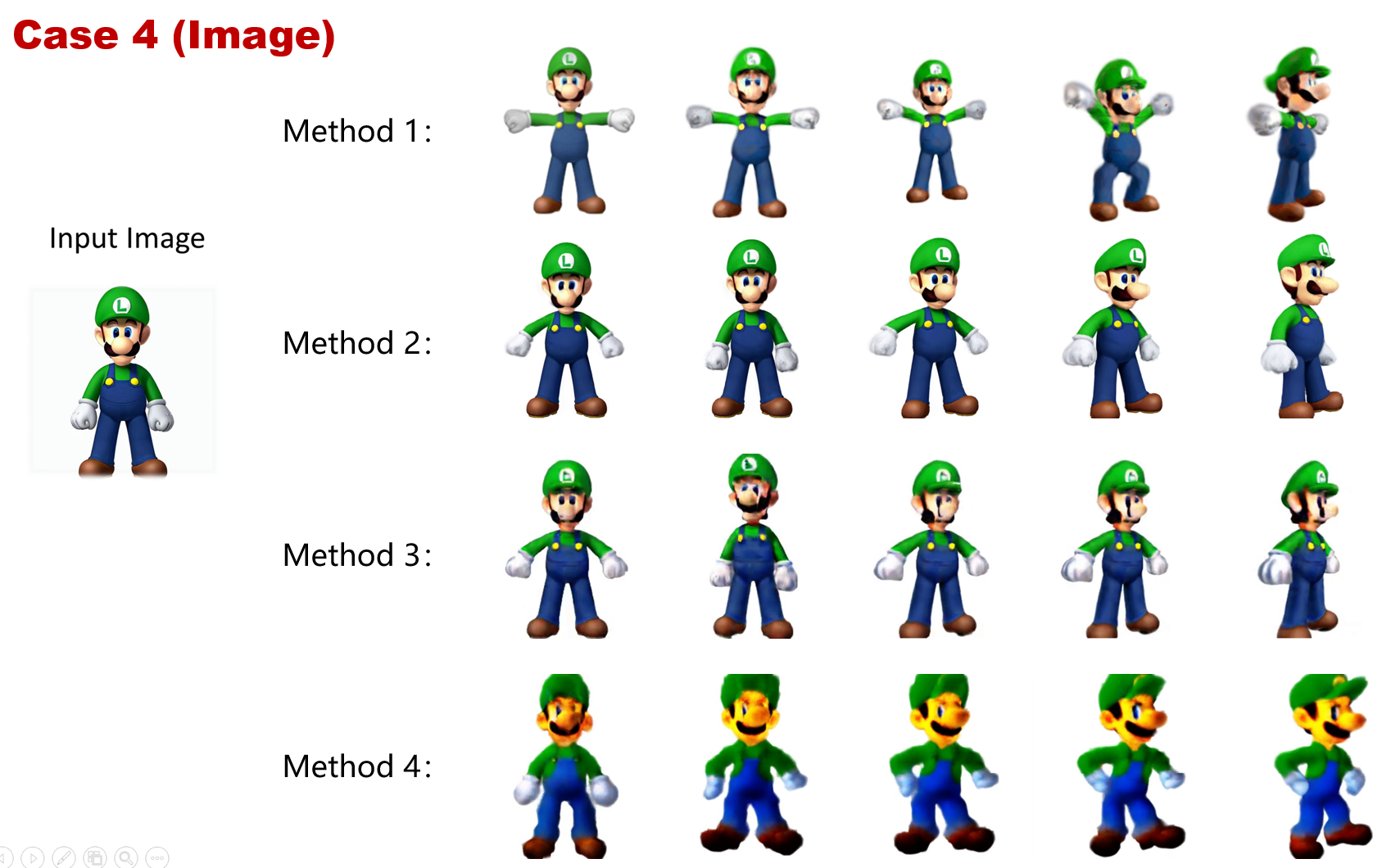}
        \caption{Screenshot of the image evaluation.}
    \end{subfigure}
    \hfill
    \begin{subfigure}[b]{0.48\textwidth}
        \centering
        \includegraphics[width=1.0\textwidth]{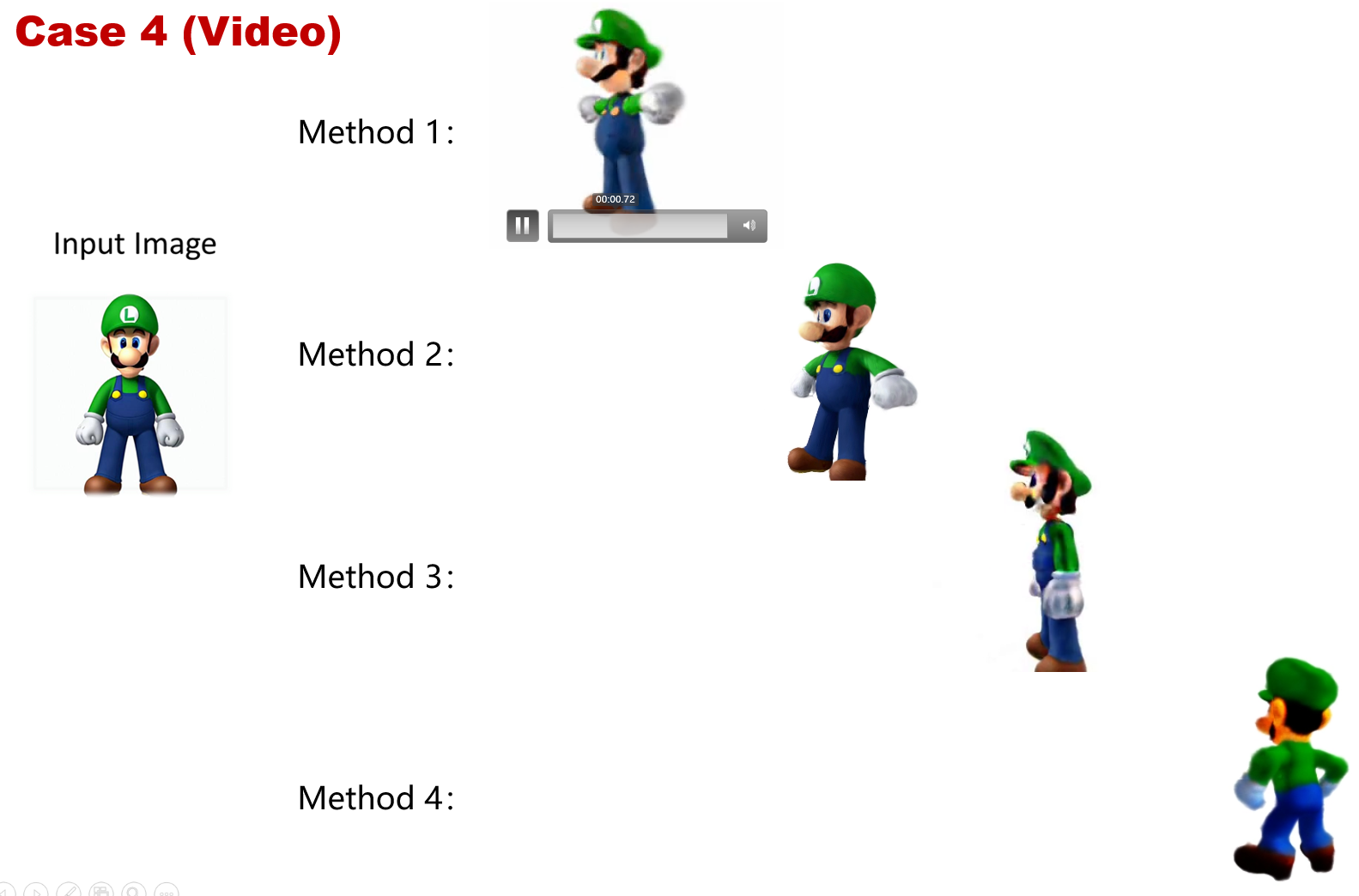}
        \caption{Screenshot of the video evaluation.}
    \end{subfigure}
    \caption{\textbf{Screenshot of our user study content.} Each participant is provided with several images and videos rendered by different methods.}
    \label{fig:screenshot_iv}
\end{figure}

We provide details of the user preference study with two screenshots. \Cref{fig:user_study_detail} illustrates the guidelines: each participant is asked to evaluate images and videos rendered by four different methods across five metrics. \Cref{fig:screenshot_iv} shows the image and video samples presented to the participants.
After comparing the images (\cref{fig:screenshot_iv}(a)) rendered by different models, participants select the method with the highest "reference image consistency" and "3D appearance". After watching the videos (\cref{fig:screenshot_iv}(b)) rendered by different models, participants select the method with the highest "motion realism" and "motion range". Finally, they choose the method with the best overall quality.
We presented several cases to 47 participants and compiled the statistics.
For statistical significance, we make the assumption of multinomial distribution, and report the 2-sigma error bar (95.6\% CI). We use standard deviation for error bar calculation.

\section{Extended Ablations}
\label{app:ablation}

\begin{figure}[t]
    \centering
    \includegraphics[width=1.0\linewidth]{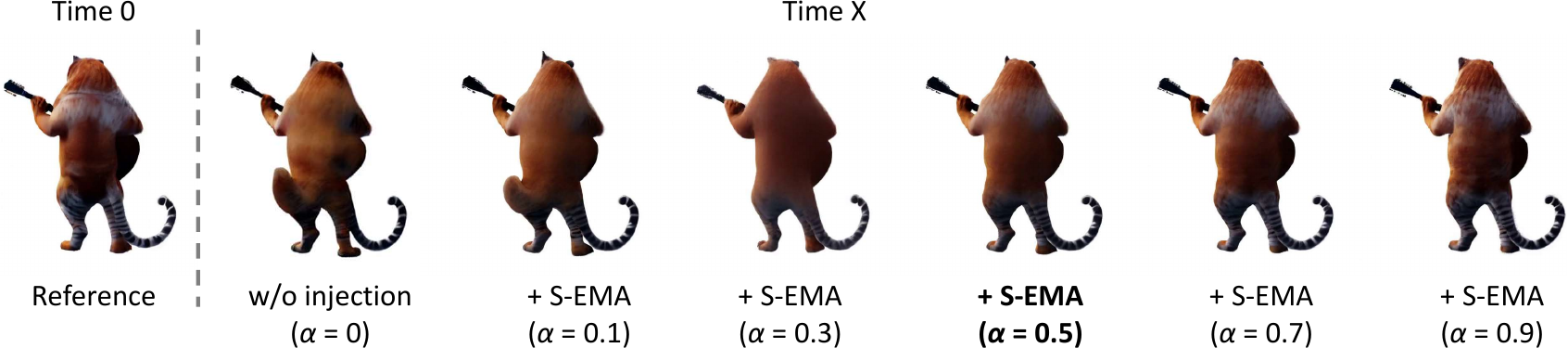}
    \caption{\textbf{Ablation on the EMA blending weight of attention injection}. 
    As the blending weight increases, the temporal consistency is significantly improved (similar white textures and consistent leg geometry).
    However, the overly high ($>0.5$) blending weight leads to a very small motion range (dynamics).
    To balance the motion range and temporal consistency, we choose the EMA weight as $\alpha=0.5$.
    Video demonstration can be found in our \emph{supplementary materials}.
    }
    \label{fig:ablation_attn_d}
\end{figure}

\paragraph{Attention injection weight.}
\cref{fig:ablation_attn_d} analyzes different EMA blending weights $\alpha$ of attention injection in the spatial attention layers.
It is obvious that the increasing blending weight benefits the temporal consistency in texture, \eg, similar white texture in the back and consistent leg geometry.
We also observe that overly high ($>0.5$) blending weight significantly attenuates the object motion range.
This trade-off can be better illustrated by the videos provided in the \emph{supplementary materials}.
Taking both motion range and temporal consistency into consideration, we choose $\alpha=0.5$ as an appropriate blending weight without sacrificing the dynamics.

\begin{wrapfigure}{r}{0.5\textwidth}
  \vspace{-10mm}
  \centering
  \resizebox{\linewidth}{!}{
  \includegraphics{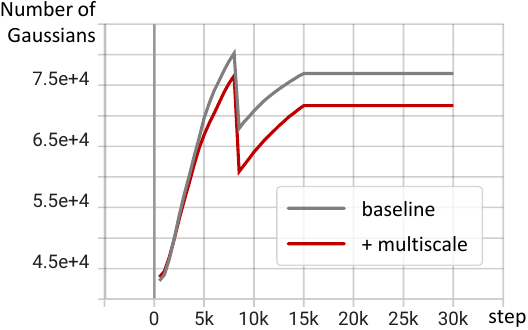}
  }
  \caption{\textbf{Additional ablation on the multiscale renderer.}
  With the multiscale rendering augmentation in Stage II (\textcolor{darkred}{darkred}), the number of Gaussians declines significantly. 
  }
  \vspace{-7mm}
  \label{fig:ms_gs}
\end{wrapfigure}

\paragraph{Number of Gaussians.}
In \cref{fig:ms_gs}, we show the number of Gaussians before and after adding the multiscale renderer.
Guo et al.~\cite{guo2024motion} observed that visual overfitting often leads to redundant Gaussian splats in dynamic scene reconstruction, which is hard to optimize and causes unsatisfying rendering results.
With the multiscale renderer, we observe a significant decline of Gaussian points, in addition to the dropped training PSNR reported in \cref{fig:ms}.

\begin{figure}[t]
    \centering
    \includegraphics[width=1.0\linewidth]{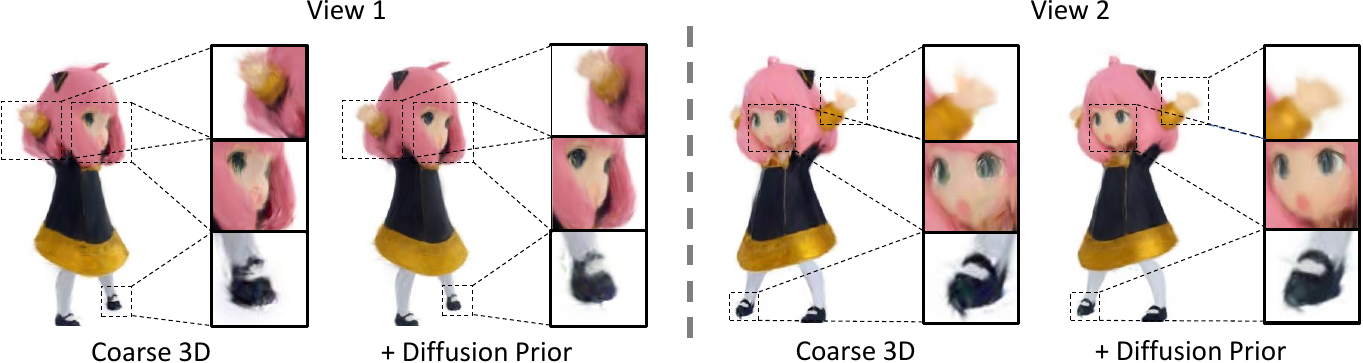}
    \caption{
    \textbf{Ablation on diffusion refinement.} 
    The left and right panels depict two different view of renderings with the case \texttt{anya}.
    The results after adding the diffusion refinement show finer facial and hand details with less noisy Gaussians.
    }
    \label{fig:ablation_refine_anya}
\end{figure}

\paragraph{Additional results for diffusion refinements.}
In \cref{fig:ablation_refine_anya}, the effectiveness of our diffusion refinement is illustrated with zoomed-in details.
It can be observed that the facial and hand details become finer and Gaussian noises are removed after the refinement stage.

\setlength{\intextsep}{0pt}

\section{Extended Results}
\label{app:res}

\paragraph{Dynamics of our results.}
For the best demonstration of our 4D model dynamics, please refer to the \emph{supplementary materials} where you can find videos generated by our 4D model.

\begin{figure}[t]
    \centering
    \includegraphics[width=1.0\linewidth]{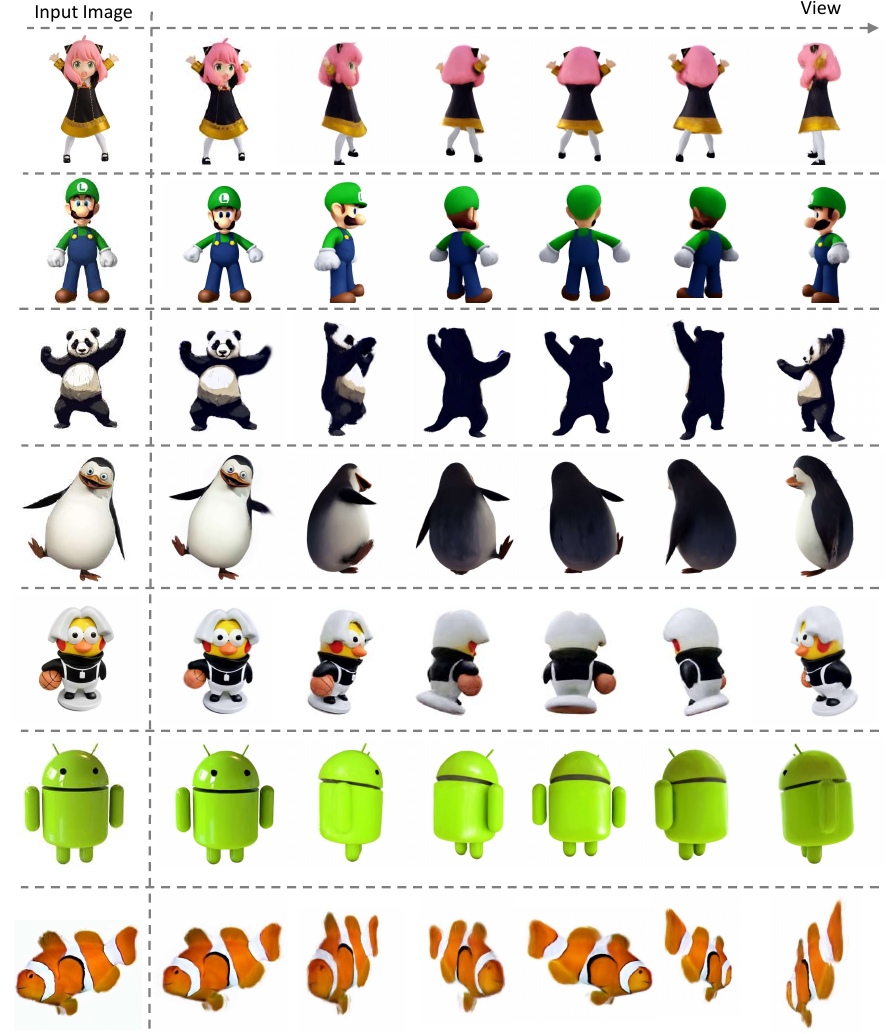}
    \caption{\textbf{Multi-view results of our models.} This figure is a supplement of \cref{fig:res} in the main paper. The 6 columns show the images rendered by our model in different views: {0$^\circ$, 90$^\circ$, 135$^\circ$, 180$^\circ$, 225$^\circ$, and 270$^\circ$}.
    Multi-view renderings demonstrate the geometry/texture consistency and promising quality of our 4D representation.}
    \label{fig:supp_mv}
\end{figure}

\paragraph{Multi-view results of our results.}
\cref{fig:supp_mv} shows the multi-view results of our 4D model, which is a supplement of \cref{fig:res}.
Due to the page limit of the main paper, we only show two views of the 4D model there, which is not enough to illustrate the 3D appearance of our model.
To this end, we render our model in more viewpoints: 0$^\circ$, 90$^\circ$, 135$^\circ$, 180$^\circ$, 225$^\circ$, and 270$^\circ$.
The rendered multi-view images show that our method can produce images with high 3D consistency and satisfactory quality.

\begin{figure}[t]
    \centering
    \includegraphics[width=1.0\linewidth]{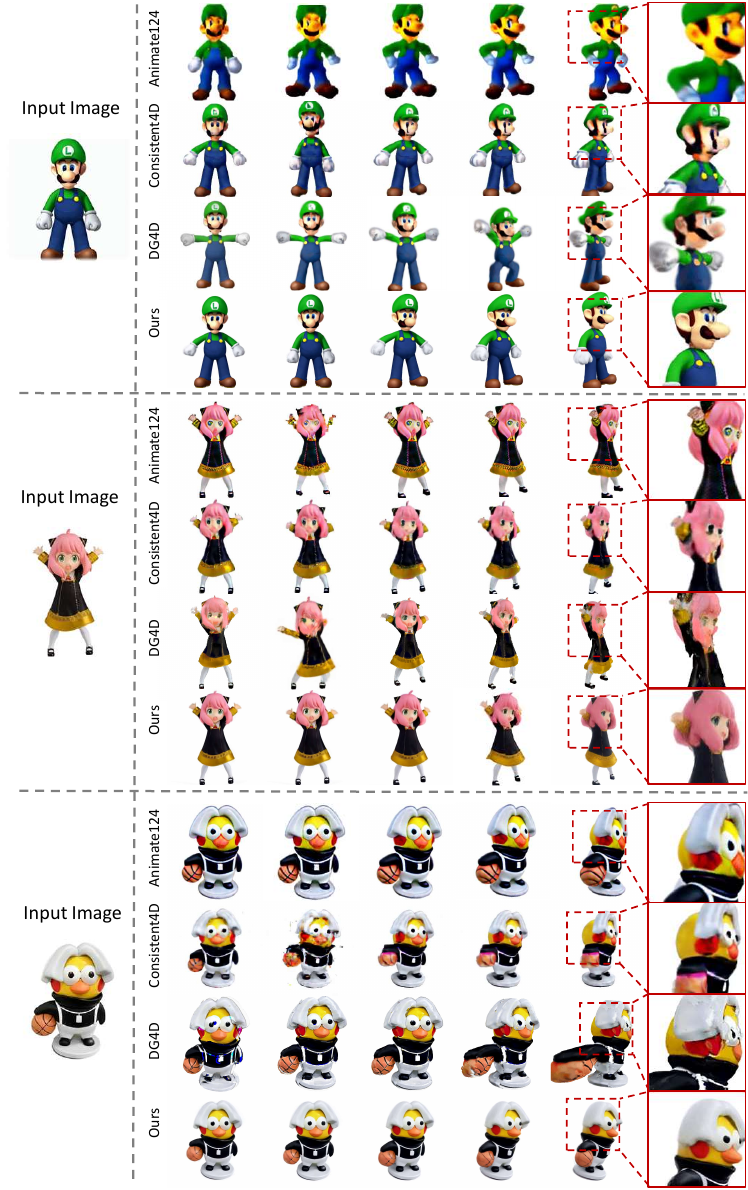}
    \caption{Comparison with Animate124~\cite{zhao2023animate124}, Consistent4D~\cite{jiang2023consistent4d}, and DreamGaussian4D (DG4D)~\cite{ren2023dreamgaussian4d} in three cases \texttt{luigi}, \texttt{anya} and \texttt{chicken-basketball} (better zoom in).}
    \label{fig:supp_comp}
\end{figure}

\paragraph{More visual comparisons.}
\cref{fig:supp_comp} provides additional visual comparisons with our baselines, continuing from \cref{fig:comp} in the main paper. We use three additional cases: \texttt{luigi}, \texttt{anya}, and \texttt{chicken-basketball}. The first two columns show animation results from the same view, while column 3 to 5 display three different views. The last column presents a zoomed-in image of the final rendered view.
Multi-view videos for visual comparison can be found in the \emph{supplementary materials}.

\paragraph{Efficiency.}
Our framework takes approximately 1 hour and 25 minutes on average for each 4D object generation. Specifically, Stage I requires about 40 minutes for video and multi-view generation; Stage II, involving 4D Gaussian Splatting optimization, takes around 25 minutes; and the refinement process takes about 40 minutes.
In previous works, Consistent4D~\cite{jiang2023consistent4d} and Animate124~\cite{zhao2023animate124} take about 2.5 and 9 hours, respectively, for 4D generation. Notably, DreamGaussian4D~\cite{ren2023dreamgaussian4d} achieves extremely short optimization time of 7 minutes.
Our optimization time falls between these, but our framework offers superior view consistency, 3D appearance, and motion quality. Since Stage I appears to be one of the efficiency bottlenecks, future work should focus on incorporating efficient sampling for video diffusion models to boost speed.

\begin{figure}[t]
    \centering
 \includegraphics[width=1.0\linewidth]{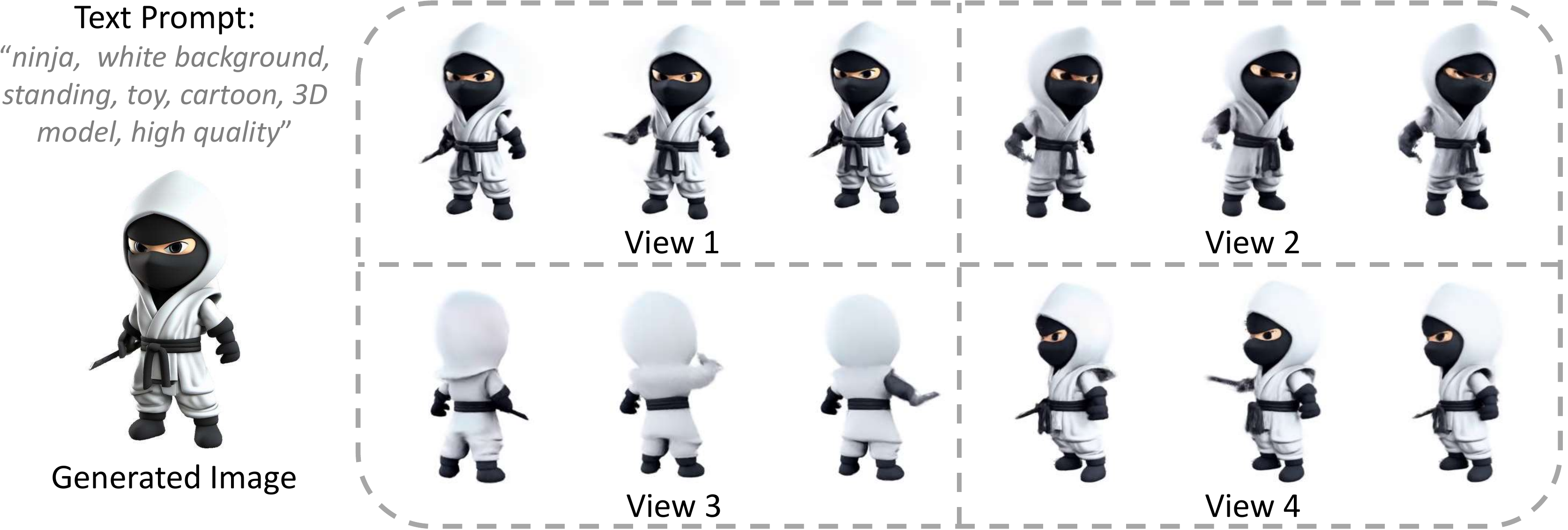}
    \caption{
    \textbf{Text-to-4D results.}
    We feed a text prompt (left top) into SDXL~\cite{podell2023sdxl} to generate a ninja image (left bottom).
    This image can be transformed into 4D objects with our framework, presenting indirect text-to-4D application.
    The right panel shows multi-view renderings of the 4D model.
    }
    \label{fig:t24d}
\end{figure}

\begin{figure}[t]
    \centering
 \includegraphics[width=1.0\linewidth]{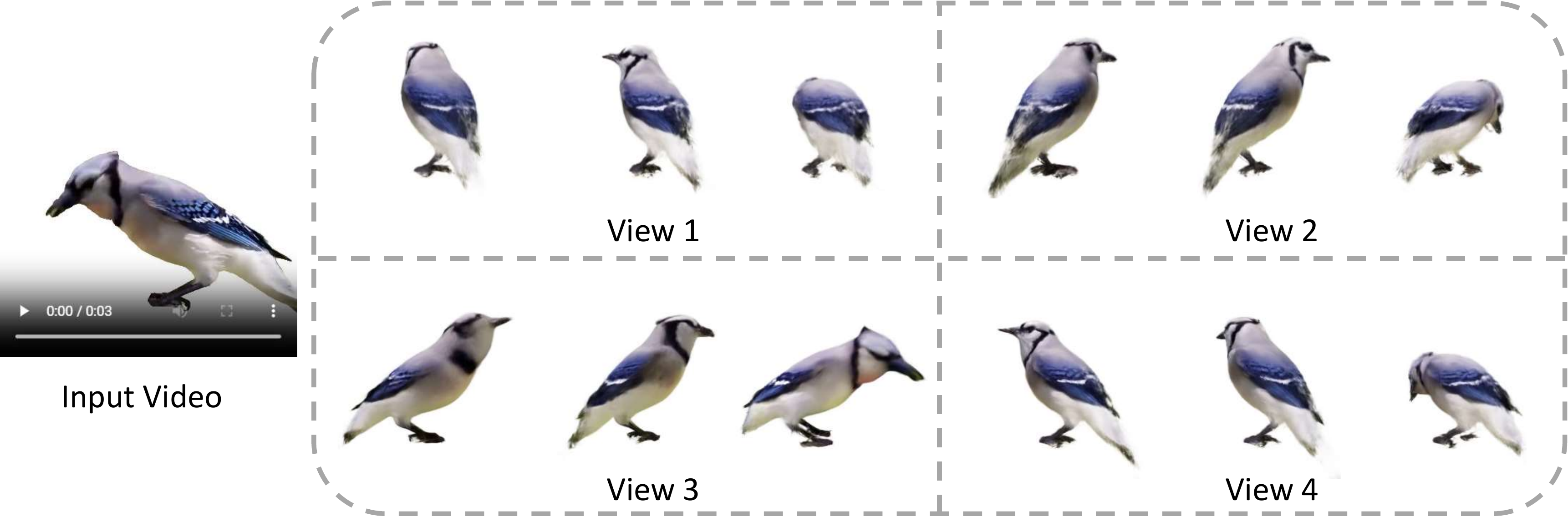}
    \caption{
    \textbf{Video-to-4D results.}
    Our framework can be seamlessly extended to video-to-4D generation.
    The right panel shows the renderings of our 4D model from four viewpoints.
    This \texttt{bird} video is taken from Consistent4D~\cite{jiang2023consistent4d}.
    }
    \label{fig:vid24d}
\end{figure}

\paragraph{More applications.}
Benefiting from our explicit generation, we can easily adapt \Ours to both text-to-4D and video-to-4D tasks.
\Cref{fig:t24d} shows the generation results of the text-to-4D. We first feed an example text prompt into SDXL~\cite{podell2023sdxl} to get the high-resolution image. Then this image is transformed into a 4D model with our framework.
\Cref{fig:vid24d} shows the results of the video-to-4D. We just skip the dynamic generation step and start with our view synthesis pipeline.



\newpage

\end{document}